# A systematic review and meta-analysis of Digital Elevation Model (DEM) fusion: pre-processing, methods and applications

Chukwuma J. Okolie[1,2*] and Julian L. Smit[1]

[1]Division of Geomatics, School of Architecture Planning and Geomatics, Faculty of Engineering & the Built Environment, University of Cape Town, South Africa

[2]Department of Surveying & Geoinformatics, Faculty of Engineering, University of Lagos, Nigeria

*Corresponding author: CJO – cokolie@unilag.edu.ng; JLS – julian.smit@uct.ac.za

**Abstract**

The remote sensing community has identified data fusion as one of the key challenging topics of the 21st century. The subject of image fusion in two-dimensional (2D) space has been covered in several published reviews. However, the special case of 2.5D/3D Digital Elevation Model (DEM) fusion has not been addressed till date. DEM fusion is a key application of data fusion in remote sensing. It takes advantage of the complementary characteristics of multi-source DEMs to deliver a more complete, accurate, and reliable elevation dataset. Although several methods for fusing DEMs have been developed, the absence of a well-rounded review has limited their proliferation among researchers and end-users. Combining knowledge from multiple studies is often required to inform a holistic perspective and guide further research. In response, this paper provides a systematic review of DEM fusion: the pre-processing workflow, methods and applications, enhanced with a meta-analysis. Through the discussion and comparative analysis, unresolved challenges and open issues are identified, and future directions for research are proposed. This review is a timely solution and an invaluable source of information for researchers within the fields of remote sensing and spatial information science, and the data fusion community at large.

## 1. Introduction

Digital Elevation Models (DEMs) are fundamental deliverables of remote sensing and photogrammetry (Schindler et al., 2011), and one of the core layers of spatial data infrastructure (SDI), national, regional, and continental topographic databases. There is growing interest in the quantitative characterisation of land surface topography to model the interplay between various earth dynamic systems, including atmospheric, geologic, geomorphic, hydrologic, and ecological processes (Wilson, 2018). As Yang et al. (2011) put it, “topography is a basic constraint and boundary condition not only to hydrologic models of flooding and runoff and atmospheric boundary layer friction theories, but also to global change and regional sustainable development research”. This emerging reality has increased the demand for wide-area DEMs such as the TerraSAR-X add-on for Digital Elevation Measurement, or TanDEM-X and global elevation data from the Global Ecosystem Dynamics Investigation (GEDI) satellite altimeter mission. This proliferation of DEMs for topographic analysis has enabled a deeper exploration of the linkages between geomorphic processes and landforms (Wilson, 2018), and for enhanced height and biomass mapping (Dubayah et al., 2020). DEMs have numerous applications in geographic information systems (GIS) and are the most common basis for digital relief maps (Elkhrachy, 2018). DEMs are integral elements for mapping, geospatial visualisation and orthophoto generation, and are also used for three-dimensional (3D) analysis in diverse areas such as hydrology, environmental modelling, urban planning and geology (Schindler et al., 2011).

The surveying and mapping communities have constantly taken advantage of emerging technologies that offer higher positioning accuracy, or that offer savings in time and cost. Consequently, the process of acquiring topographic data over the past decades has seen the mapping industry move from classical ground survey methods to passive and active sensing approaches. The emergence of remote sensing technologies have not only improved the speed of data acquisition but have also provided elevation data for areas that are difficult to access and survey (d’Ozouville et al., 2008; Fuss, 2013). The acquisition modes and processing techniques used in the production of DEMs have been modernised. Some of these techniques include: short-range 3D cameras (e.g., Mankoff and Russo, 2013), digital photogrammetry (e.g., Smit, 1997; Onwudinjo and Smit, 2019; Iheaturu et al., 2020; Gbopa et al., 2021), synthetic aperture radar interferometry, or InSAR (e.g., Avtar et al., 2015; Rizzoli et al., 2017), radargrammetry (e.g.,

Agrawal et al., 2018), and light detection and ranging, also known as LiDAR (e.g., Dubayah et al., 2020).

The remote sensing platforms, sensors and DEM reconstruction algorithms in use today have made wide-area DEMs more readily available. These DEMs have different characteristics in terms of spatial coverage, resolution properties, and accuracy levels (Petrasova et al., 2017). The differences in sensor characteristics have led to an abundance of DEM products with varying qualities and applications. Also, there are differences in the type of surface model (e.g., bare-earth or above-ground), the spatial distribution of errors, the vertical datum and first- or second-order trends (Reuter et al., 2007). To take advantage of the different DEM products in use today, researchers have resorted to fusion for deriving more comprehensive and enhanced digital elevation datasets that combine the complementary characteristics of source DEMs.

The remote sensing community has identified data fusion as one of the key challenging topics of the 21st century. This realisation has reflected in the scientific vision and research agenda of professional societies such as the International Society for Photogrammetry and Remote Sensing (ISPRS, Chen et al., 2016) and the IEEE Geoscience and Remote Sensing Society (IEEE GRSS Strategic Plan, 2020). The aim of remote sensing data fusion is the integration of information acquired with different resolution characteristics (e.g., spatial or spectral) from sensors mounted on platforms (e.g., aerial, satellite or ground) to produce fused data that is more detailed than each of the sources (Zhang, 2010). There has been tremendous growth in the scientific literature on remote sensing data fusion, and this reflects the increased role of data fusion at the core of many applications in remote sensing. A significant part of the scientific output in remote sensing data fusion has been focused on two-dimensional (2D) image fusion, either for classification purposes or for pansharpening (Schmitt and Zhu, 2016). Van Genderen and Pohl (1994) define image fusion as "the combination of two or more different images to form a new image by using a certain algorithm". Satellites in orbit continue to acquire data in different parts of the electromagnetic spectrum leading to various satellite image products such as panchromatic (PAN), hyperspectral (HS), multispectral (MS), and synthetic aperture radar (SAR) imagery which convey different information about the areas on earth under observation (Kulkarni and Rege, 2020). The fusion of remotely sensed images is an effective approach for generating a value-added image that has more comprehensive spatial and spectral information, and surface information (Ghassemian, 2016; Kulkarni and Rege, 2020).

We define DEM fusion as the synergistic integration of two or more source DEMs using a certain algorithm to generate an improved DEM which is more detailed than each of the sources. The aim is to generate a DEM of higher quality. However, the definition of quality would depend on the application. DEM fusion has received little attention unlike several other “hot” topics in remote sensing image fusion (e.g., MS-PAN and MS-HS image fusion). The absence of a well-rounded and comprehensive review on the subject has limited the adequacy of knowledge for researchers. This is complicated by the fact that the available studies on DEM fusion are often heterogeneous in their research design, data sources and processing workflows. Given the foregoing, it is desirable to carry out a comprehensive systematic review and meta-analysis on DEM fusion. This is a feasible solution for aggregating the available body of knowledge on the subject and for updating the remote sensing and allied fields on recent advances. This review is aimed at addressing the following questions: *(i) what are the characteristics and relative performances of the various DEM fusion methods, including the strengths and weaknesses? (ii) what are the relevant pre-processing operations required for DEM fusion, (iii) what are the effects of DEM properties on the quality of fusion achieved, and (iv) what are the effects of terrain attributes/parameters on the fusion quality?* The main contributions of the paper are summarised below:

i. Description of the general approach and workflow of DEM fusion, including the pre-processing phase.
ii. A comparative review of the characteristics of the investigated methods, performance evaluation in terms of strengths and weaknesses, and analysis of relevant parameters that influence fusion results.
iii. Based on the in-depth analysis and review, unresolved issues and prospects for future research are presented, which are valuable for practitioners and researchers in the field of remote sensing and spatial information sciences, and the data fusion community at large.

To our knowledge, this is the first systematic review of DEM fusion methods. The remainder of this paper is organised as follows: an overview of DEM fusion including the general approach and workflow is given in section 2 while the methodology adopted for this review is presented in section 3. Section 4 presents the results and discussion, including the pre-processing workflow, review of the DEM fusion methods, pros and cons, analysis of DEM sources, parameters influencing weight maps and height error maps used in fusion, accuracy assessment and applications of DEM fusion. The research challenges, open issues, and future directions are presented in section 5.

## 2. General Workflow of DEM Fusion

DEM fusion is an important application case of data fusion in remote sensing (Schmitt and Zhu, 2016; Bagheri et al., 2017). It stems from ongoing developments in remote sensing image fusion, and mostly occurs at the pixel level. According to Papasaika et al. (2008), the fusion of overlapping DEMs acquired via different techniques or at different periods enables the detection of inconsistencies, eliminates gaps in the data coverage and leads to improvement of the density and accuracy of the datasets. Since DEMs differ in their geometric characteristics, accuracy levels, spatial resolution, temporal resolution, and coverage, data fusion helps to combine the advantages of the different data sources to provide a value-added product that is more complete, accurate and reliable (Karkee et al., 2008). The practice of fusing multi-source and multi-sensor DEMs with variable spatial coverages has been well reported in the literature, and several techniques have been developed and applied by researchers over the years. DEM fusion can be viewed as an ill-posed problem. A problem is said to be ill-posed in the following situations: if the solution to the problem is not unique; if the problem does not have a solution; or, if the result can be significantly altered by a small change in the input data (Hadamard, 1902; Beyerer et al., 2008; Sadeq et al., 2016).

There are several situations where the fusion of DEMs is required. For example, for merging of DEMs generated with different technologies, updating DEMs with more recent ones, enhancing DEMs using available auxiliary elevation data from other DEMs, improvement of DEM accuracy through the removal of systematic errors, and fusion of InSAR DEMs with varying acquisition geometries. The integration of mutually independent DEMs can lead to much more semantic information about terrain geomorphology (Podobnikar, 2005). The goal in DEM fusion is to achieve a continuous, contiguous and complete representation of the terrain. For a detailed explanation of situations requiring DEM fusion, see Buckley and Mitchell (2004).

From a remote sensing perspective, Schmitt and Zhu (2016) categorised the data fusion workflow into four tasks: data alignment, data/object correlation, attribute estimation, and identity estimation, and further summarised these tasks into two core actions: matching and co-registration (data alignment and data/object correlation) and the actual fusion process (attribute/identity estimation). With respect to image fusion, a general workflow consisting of data alignment (radiometric and geometric correction) followed by the actual fusion process was presented by Pohl and van Genderen (2016). Before the implementation of image fusion, some key issues need to be addressed by the user. These questions include: the objective/application, the most relevant datasets for meeting the needs, the 'best' technique for fusing the data for that application, the required pre-processing steps and the combination of data that would meet the need (Pohl and van Genderen, 1998). Following the review of studies (discussed in section 4), we have generalised the DEM fusion procedure into five key stages shown in Figure 1 and discussed below.

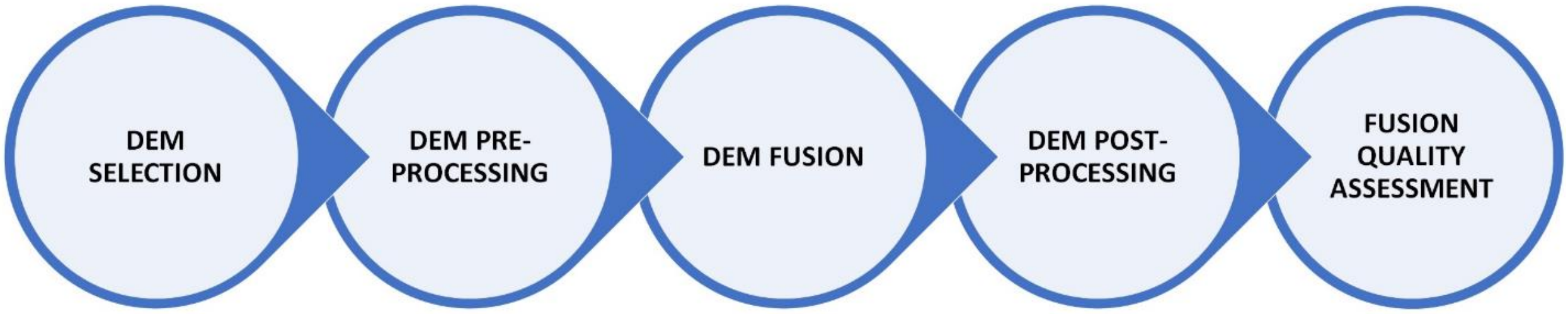

Figure 1: The generalised procedure for DEM fusion starts with the selection of source DEMs followed by pre-processing, the actual fusion, post-processing and quality assessment of the fused DEM

## 2.1 DEM selection

The choice of the source data depends on the purpose of the fusion, and the desired application (Pohl and van Genderen, 2016). It very much depends on data availability, data quality, sensor characteristics, and availability/accessibility of effective algorithms for extracting information (Salentinig and Gamba, 2015; Pohl and van Genderen, 2016). A primary consideration when selecting DEMs is to consider their complementary characteristics. For example, the selection can be done in such a way that the limitations of one sensor (e.g., SAR layover and shadow) are compensated for by the other sensor (e.g., LiDAR). Another approach is to combine data produced from the same sensor but with different sensor geometries or configurations (e.g., the combination of ascending and descending pass SAR). The optimum selection of data can be hindered by several constraints including finance, data access restrictions, availability of suitable software and computational resources, availability of data storage space or facilities, available internet resources, and data latency. Data quality issues and atmospheric effects (e.g., cloud cover) can also hinder data selection. The global outbreak of the COVID-19 pandemic in early 2020 has also proven that global emergencies can hinder optimum data selection. To curtail the spread of the disease, several countries enforced lockdowns, and this prevented many researchers from accessing imageries and spatial datasets stored in hard drives/servers at offices and laboratories. An initial accuracy assessment can also guide the selection of the most suitable DEMs to use. The selection of data to partake in image fusion is closely followed by pre-processing.

### 2.2 DEM pre-processing

Before fusion, issues that compromise the elevation values recorded by the remote sensor should be eliminated (Pohl and van Genderen, 2016). There are a variety of factors that affect the quality of source DEMs. These include: the data collection techniques, spatial scales, noises etc. (Chaplot et al. 2006; Chen and Yue 2010; Yue et al., 2015). Factors such as the spatial resolutions, registration errors, void/invalid pixels, relative vertical discrepancies and horizontal displacements among the multi-source datasets should be taken into consideration (Fisher and Tate 2006; Yue et al., 2015). The sources of errors and differences between multi-source DEMs have been treated extensively by Buckley and Mitchell (2004). Another issue is that the geodetic datums, coordinate systems and formats of DEMs tend to be inconsistent (Fu and Tsay, 2016). Pre-processing is required to harmonise the disparities in spatial resolutions or grid sizes. Some of these disparities include the differences in grid cell size and coordinates which could lead to discrepancies between DEMs (Bastin et al., 2002; Van Niel et al., 2008; Pham et al., 2018). Common pre-processing operations include vertical accuracy improvement, void filling, co-registration, resampling, and generation of height error maps (HEMs) or weight maps.

### 2.3 DEM fusion

DEM fusion usually takes place at the pixel level. Several techniques have been applied by researchers over the years and different nomenclatures have been adopted in their descriptions. A common feature of many methods is the use of height error maps (HEMs) or weight maps to determine the influence of each source DEM in the fusion. The fusion methods are discussed in section 4.

### 2.4 DEM post-processing

The fused DEM can be subjected to further processing, to smoothen the derived elevation surface, blend zones or transition areas, through geomorphologic enhancements with the use of trend surfaces, de-noising algorithms and spatial filters (e.g., adaptive mean and gaussian filters, high pass filters). An application-driven point reduction filter can also be applied for optimising the fused DEM to derive a more efficient point spacing (Buckley and Michell, 2004).

### 2.5 Fusion quality assessment

In the literature, three approaches are commonly used for assessing the quality of fused DEMs. The first is a qualitative approach in which the user does a visual inspection and comparison of the results. This could also involve the analysis of height profiles or terrain attributes generated from the fused DEM such as hillshade, and slope maps (e.g., Sadeq et al., 2016). The second is a quantitative approach in which the user employs statistical metrics to measure the relative or absolute quality of the fused DEM. Relative quality is the closeness of the fused DEM to the original DEM while absolute quality is the similarity between the fused DEM and a reference DEM. In the third approach, the concern is more about the performance of the fused DEM in certain contexts or applications e.g., modelling surface flow, land cover mapping and classification, modelling volcanic flow, delineating watersheds, and not necessarily with the per-point accuracy statistics.

## 3. Review Methodology

For the systematic search and review procedure, we followed the Preferred Reporting Items for Systematic Reviews and Meta-Analysis guidelines - PRISMA 2020 (Page et al., 2021a,b; http://www.prisma-statement.org/). Essentially, the workflow is divided into three stages: identification, screening and inclusion. The PRISMA flow diagram for this review is shown in Figure 2, and the stages are explained in the sections that follow.

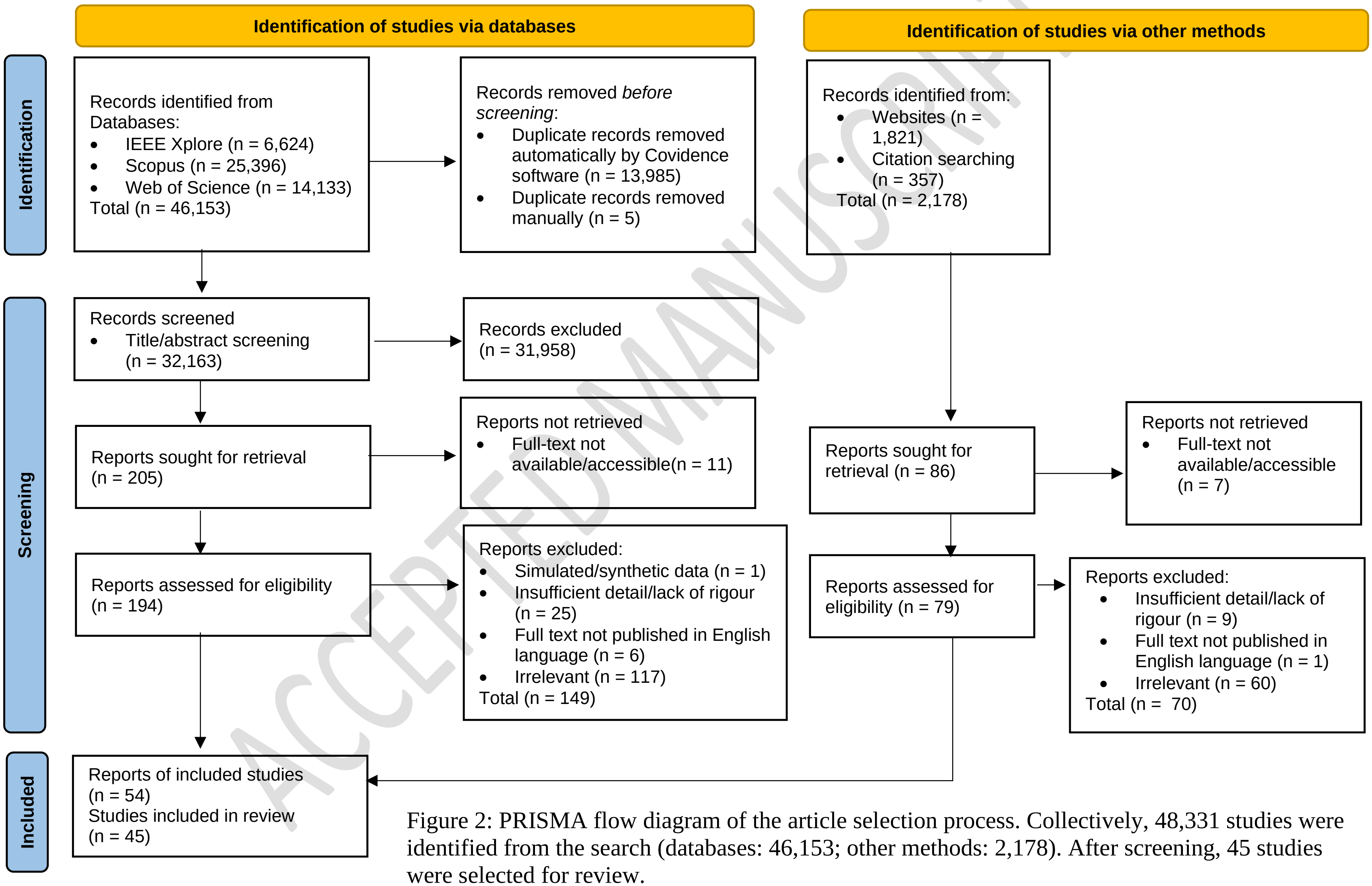

Figure 2: PRISMA flow diagram of the article selection process. Collectively, 48,331 studies were identified from the search (databases: 46,153; other methods: 2,178). After screening, 45 studies were selected for review.

## 3.1 Eligibility criteria

To select relevant articles, the following criteria were applied:

### *3.1.1 Inclusions*

The following studies were included:

1. Methods or algorithms for DEM fusion with a sound theoretical basis. The method should be based on a well-founded and logical theoretical framework.
2. Fusion of grid/raster-based DEMs. Regular matrices or uniform grids are the standard format in which DEMs are stored.
3. Fusion of multi-source and multi-sensor DEMs. DEMs are acquired from various sources, and are produced from a multitude of sensors that have varying degrees of sensitivity in the electromagnetic spectrum.
4. Studies applying fusion algorithms to fuse multi-source DEMs along blend/transition zones. Fusion can be applied across the entire DEM coverage area or specifically along the transition zones between DEMs.
5. Fusion experiments using real-world earth data. Unlike synthetic data, real-world datasets provide more realistic perspectives on the performance of the fusion methods.
6. Advanced geostatistical methods developed for DEM fusion. While many traditional geostatistical methods are usually used for interpolating scattered elevation data, some advanced methods that satisfy the premise of fusion have been developed.
7. The articles incorporated either a quantitative or qualitative assessment of the fusion quality. This is necessary to understand the merits and limitations of each method.

### *3.1.2 Exclusions*

The following studies were excluded.

1. Studies in which the DEM fusion is treated abstractly or in which the analysis presented is purely conceptual.
2. Studies dealing with top-of-canopy DEMs, or canopy height models (CHMs), without consideration for ground-level topography.
3. Studies dealing with fusion or conflation of digital terrain models (DTMs), vector hydrographic data or triangulated irregular networks (TINs)/triangulated meshes. Besides the values of height, a DTM consists of other topographic attributes or terrain morphological elements such as slope, aspect, curvature, gradient, skeleton (sinks, valleys, saddles, ridges, and peaks) etc. (Podobnikar, 2005).

4. Studies that combine representations from various DEMs but without an actual fusion.
5. Studies that demonstrate fusion using data covering extra-terrestrial bodies.
6. Studies on the fusion of bathymetric elevation datasets, or topography-bathymetry (topo-bathy) fusion.
7. Studies that do not fuse DEMs but that employ a fusion of methods such as stereo-SAR and InSAR to derive a DEM.
8. Studies that lacked sufficient rigour in their approach.
9. Studies on the following topics were excluded for either of two reasons: (i) they did not satisfy the premise of DEM fusion in this work, or (ii) the existing studies on the topics are of sufficient quantity to be covered in other reviews.
    a. Studies on void filling of DEMs. Void filling algorithms are generally used to plug in gaps, holes or sinks within DEMs using auxiliary elevation data.
    b. Traditional mosaicking of DEM tiles to increase coverage area.
10. Articles published in potentially predatory journals and conferences.
11. Due to limited facilities for language translation, articles published in languages other than English were excluded.

### 3.2 Identification of studies

The search for studies was restricted to journal and conference papers. In the main search, title/abstract/keyword searches were conducted in 3 scholarly databases (Scopus, Web of Science core collection, and IEEE Xplore) without any limit on the period of coverage. The main keywords are ‘Digital Elevation Model’ and ‘Fusion’. The search query was built using Boolean operators (AND/OR) to combine related terms. A total of 46,153 search results were identified from the databases. Another 2,178 studies were identified using two additional methods (citation searching and websites search). To check for citations, we searched the in-text citations and reference lists of studies from the databases that were deemed eligible. In the websites search, we explored the Connected Papers database (https://www.connectedpapers.com/) to generate graphs and records of papers connected to the studies deemed eligible from the database search. Connected Papers is linked to the Semantic Scholar Paper Corpus (https://www.semanticscholar.org/), a free AI-powered research tool for scientific literature. It enabled the identification of additional papers (including prior works and derivative works). All the downloaded studies were imported into Covidence software for screening. Table 1 shows the different keyword combinations, the search query used and the search dates.

Table 1: Keywords used, search query string and search dates

<table>
<tr><th rowspan="2">Primary keywords</th><th colspan="2">Additional/alternative keywords</th></tr>
<tr><th>Digital Elevation Model</th><th>Fusion</th></tr>
<tr><td>1. Digital Elevation Model<br>2. Fusion</td><td>1. Digital Terrain Model<br>2. Digital Surface Model<br>3. Digital Elevation Data<br>4. DEM<br>5. DTM<br>6. DSM<br>7. Multi-source Elevation Data<br>8. Multi-sensor Elevation Data<br>9. SAR<br>10. LiDAR</td><td>1. Integration<br>2. Combination<br>3. Merging<br>4. Blending<br>5. Fusing</td></tr>
<tr><th colspan="3">Search query string</th></tr>
<tr><td colspan="3">(("Digital Elevation Model" OR "Digital Terrain Model" OR "Digital Surface Model" OR "Digital Elevation Data" OR "DEM" OR "DTM" OR "DSM" OR "Multi-source Elevation Data" OR "Multi-sensor Elevation Data" OR "SAR" OR "LiDAR") AND ("Fusion" OR "Integration" OR "Combination" OR "Merging" OR "Blending" OR "Fusing"))</td></tr>
<tr><th colspan="3">Search dates</th></tr>
<tr><td colspan="3">Databases:<br>• IEEE Xplore: 6th November 2020<br>• Scopus: 7th November 2020<br>• Web of Science (WoS) core collection: 19th November 2020<br>Website:<br>• Connectedpapers.com: 14th June 2021</td></tr>
</table>

### 3.3 Article screening

A double screening strategy (Page et al., 2021b) was adopted to help minimise screening bias and for the resolution of inclusion/exclusion conflicts through discussion and consensus. The screening was mainly done within the Covidence software environment. Covidence is a web-based software that facilitates the systematic review workflow (Veritas Health Innovation, 2021). In Covidence, thousands of articles that did not satisfy eligibility criteria were eliminated. Covidence has the advantage of automatic detection and removal of duplicate articles. For example, 13,985 duplicate records from the databases were automatically removed by Covidence, while 5 duplicate records were removed manually. After the removal of duplicates, the titles/abstracts of the remaining 32,163 articles were screened manually. The sequence of steps in the article screening is shown in Figure 2.

### 3.4 Data extraction and analysis of included studies

A total of 54 reports (29 journal papers and 25 conference papers) were deemed eligible after screening. Some studies were featured in two reports; thus 45 studies (28 journal papers and 17 conference papers) were identified from the reports. The following data were extracted from these included studies for analysis: article title, publication date, article type, list of authors, fusion method, country/region of study, pre-processing techniques, source DEM type and spatial resolution, reference DEM type and spatial resolution, source and reference DEM production technologies, study area land cover and terrain class. These data were grouped into different categories and synthesised for presentation.

## 4. Results and Discussion

### 4.1 Characteristics of reviewed studies

Figure 3 shows the word cloud for this review. The expressions, ‘Digital Elevation Model’ and ‘Fusion’ which were specified in the article search are very prominent in the word cloud. The most frequently occurring DEM product is the TanDEM-X. Other prominent DEMs include the Shuttle Radar Topography Mission (SRTM) DEM, the Advanced Spaceborne Thermal Emission and Reflection Radiometer (ASTER) Global DEM and Cartosat-1 DEM. InSAR appears as the most common DEM production technology followed by LiDAR/laser altimetry. In terms of the fusion method, weight maps which are commonly used in weighted average fusion, and variational methods are well featured.

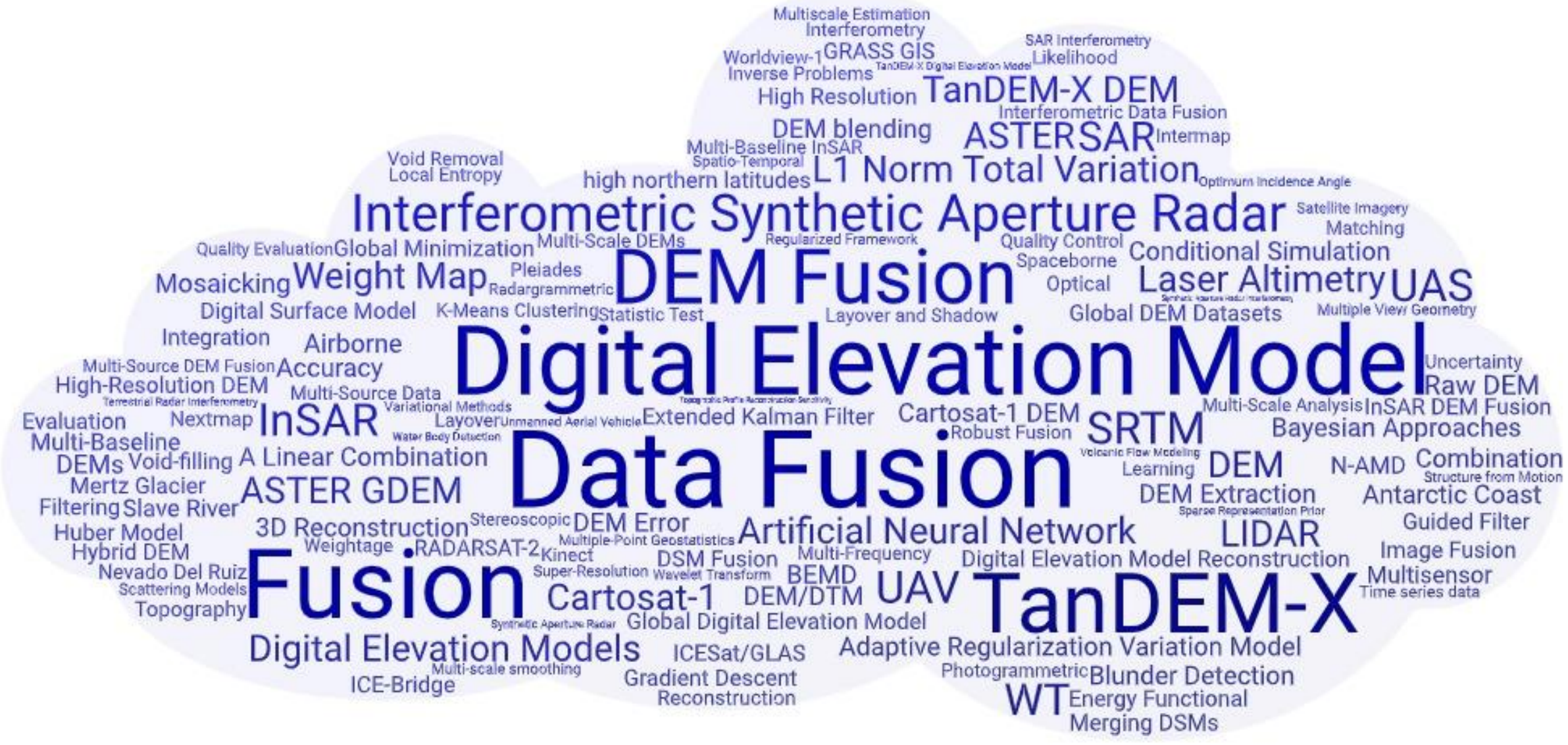

Figure 3: Word cloud showing prominent keywords from the included studies. The sizes of the font are directly proportional to the frequency of occurrence of the terms shown.

The trend in the number of publications shown in Figure 4 has been quite irregular with no steady growth. However, 53% of the total number of studies were published between 2014 and 2018. This period closely coincides with the data acquisition period of the twin SAR satellites, TanDEM-X and TerraSAR-X between 2010 and 2015, with data processing completed in 2016 (Rizzoli et al., 2017). DEM fusion has not been trending in the remote sensing data fusion community unlike other “hot” topics such as pansharpening and classification, and application areas such as change detection. For example, it has not featured in any of the data fusion contests organised by the Image Analysis and Data Fusion Technical Committee (IADF TC) of the IEEE Geoscience and Remote Sensing Society (IEEE-GRSS) since 2006. The analysis also shows a very strong disciplinary linkage as most of the DEM fusion papers are domiciled in remote sensing journals (e.g., IEEE Journal of Selected Topics in Applied Earth Observations and Remote Sensing; IEEE Transactions on Geoscience and Remote Sensing; ISPRS Journal of Photogrammetry and Remote Sensing; and Remote Sensing of Environment) and conference proceedings from remote sensing professional societies (e.g. ISPRS International Archives of the Photogrammetry, Remote Sensing and Spatial Information Sciences; and the IEEE International Geoscience and Remote Sensing Symposium). However, fusion-specific journals have since emerged (e.g., Information Fusion; and International Journal of Image and Data Fusion) thus offering more competitive publication outlets.

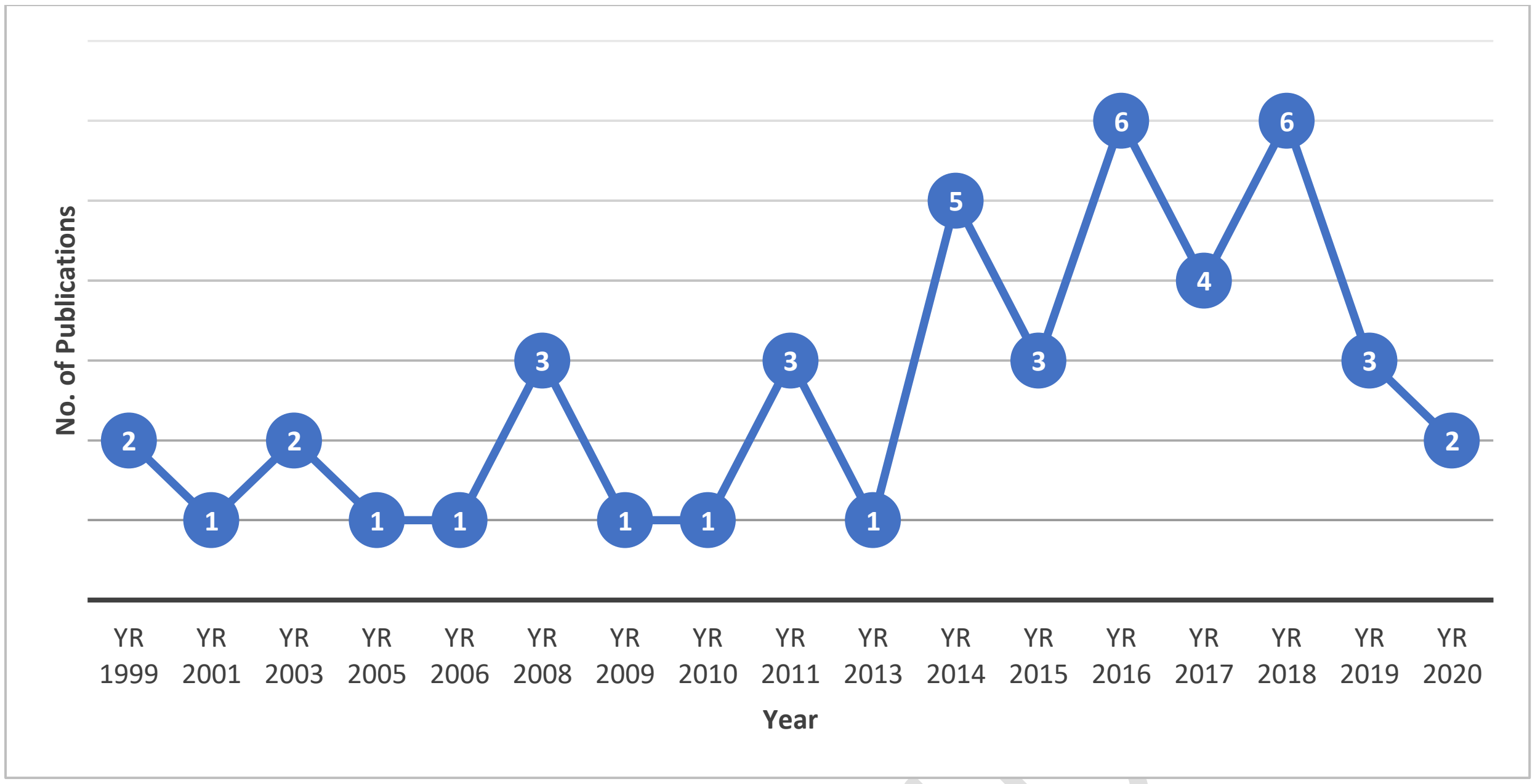

Figure 4: Trend of published papers in DEM fusion from 1999 to 2020. The highest number of publications were recorded in 2016 and 2018.

Figure 5 presents a world map showing the geographic distribution of study sites covered in the included studies. The countries with the highest number of sites are the United States (11), China (7), Germany (5), Switzerland (5) and Spain (4). The most prominent research institutions in these countries are the remote sensing laboratories at the Signal Processing in Earth Observation (Technical University of Munich) and the Remote Sensing Technology Institute (German Aerospace Centre) both in Germany; and the State Key Laboratory of Information Engineering in Surveying, Mapping and Remote Sensing at Wuhan University, China. There is a clear global north dominance in the distribution with hardly any studies emanating from global south countries in South America and Africa. Although Colombia and Djibouti had 1 site each, the studies were led by researchers from other institutions in more advanced countries. This points to a skills gap and poor utilisation of earth observation (EO) data by researchers in the global south.

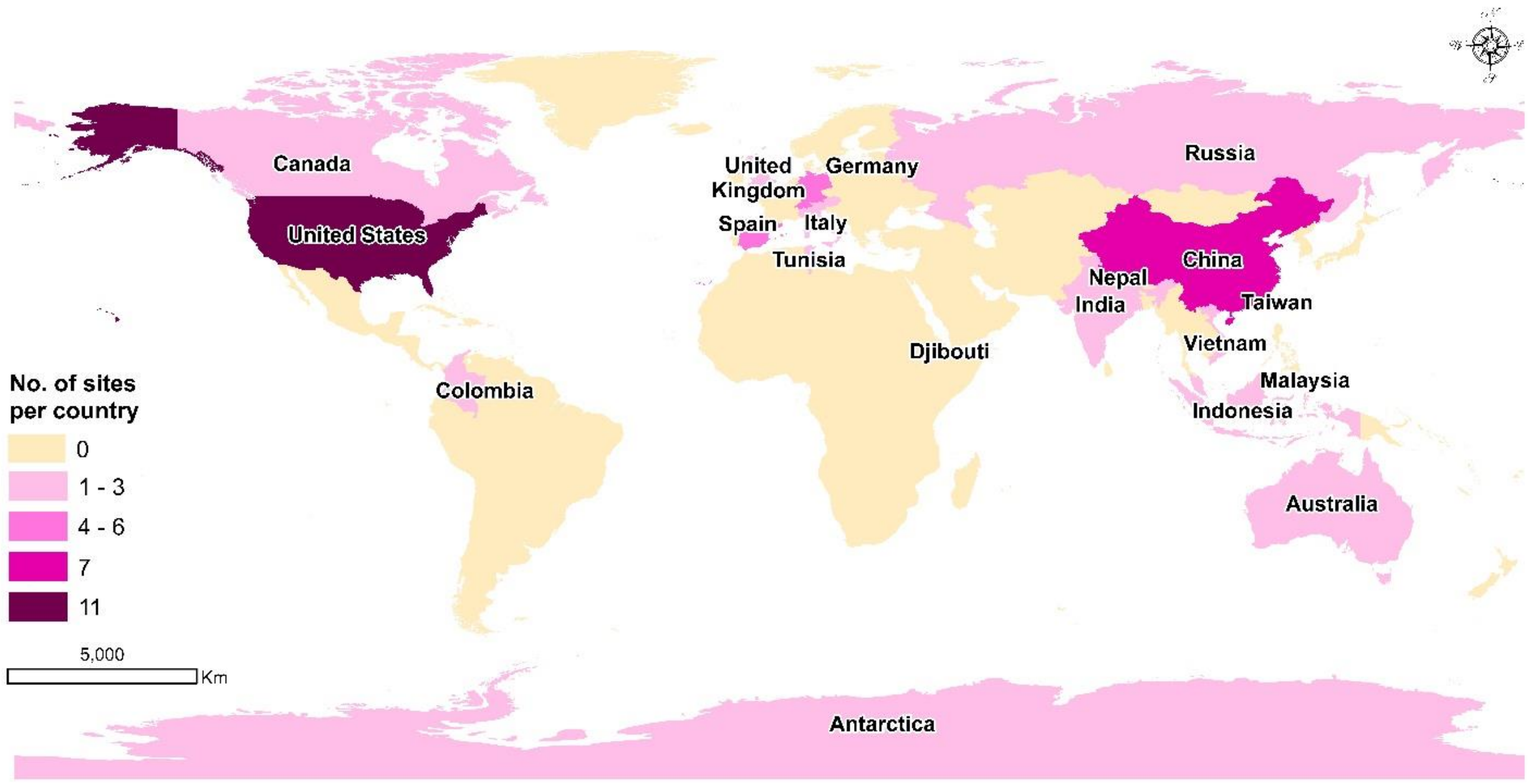

Figure 5: World map showing the geographic distribution of sites covered in the included studies. In summary, 27% of the sites are in Asia, 36% in Europe, and 25% in North America. The less featured continents include Africa (4%), Australia (4%), South America (2%) and Antarctica (2%)

## 4.2 DEM pre-processing workflow

We present in this section, a general workflow for DEM pre-processing prior fusion (illustrated graphically in Figure 6). The exact pre-processing steps to adopt are dependent on factors such as the quality and characteristics of the source DEMs. The stages in the workflow are discussed below.

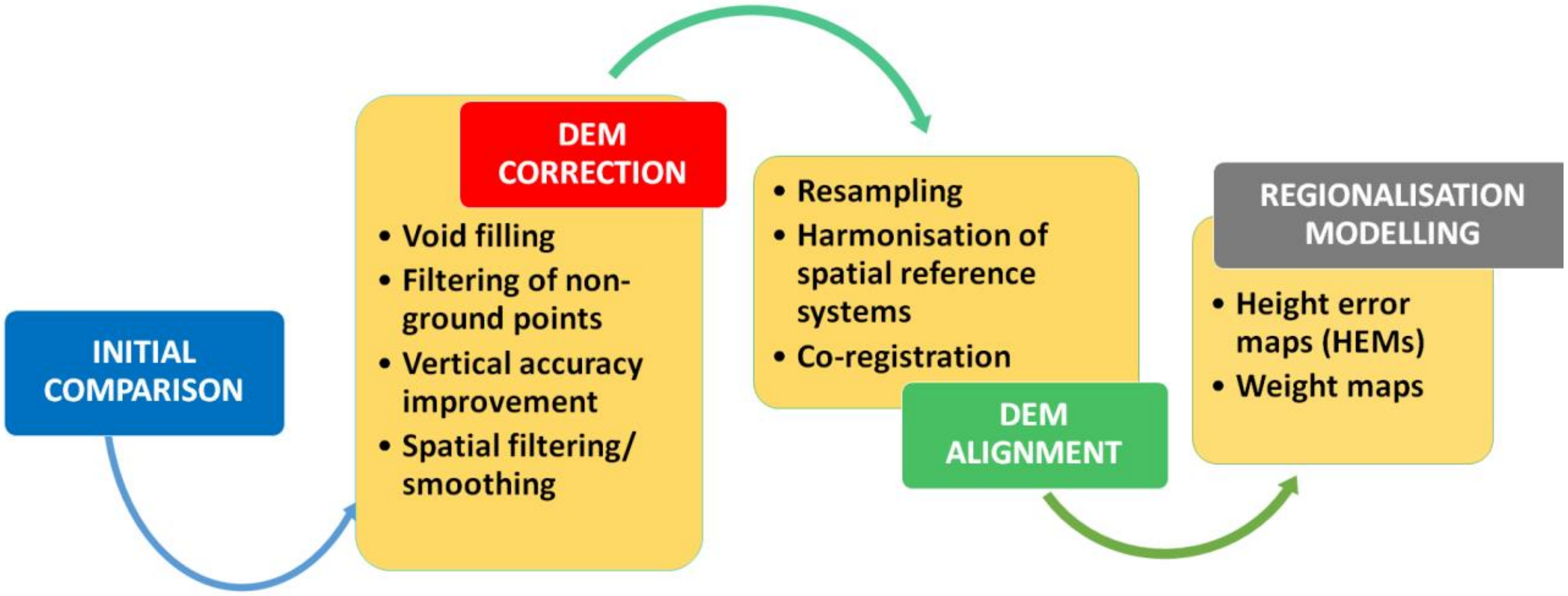

Figure 6: A generalised workflow showing the stages in DEM pre-processing. The quality of the source DEMs are improved in the second stage through correction. The third stage involves the alignment of source DEMs through resampling and co-registration to achieve coincident pixels, and harmonisation of spatial reference systems.

### *4.2.1 **Initial comparison***

Before fusion, the source DEMs will offer separate, unrelated perspectives of the same environment, with each DEM being affected by deviations from the 'true' surface, and the first task is to calculate the discrepancies between the DEMs through statistical analysis and DEM differencing (Buckley and Mitchell, 2004).

### *4.2.2 **DEM correction***

#### *4.2.2.1 Void filling*

Voids are a serious deficiency that impacts the quality, completeness, and utility of DEMs. Voids constitute systematic error and occur for a variety of reasons in different terrain types. For example, voids due to complex dielectric constant are more likely to occur in desert areas while voids caused by shadowing are common in mountainous areas (Reuter et al., 2007). The frequency of voids with respect to elevation has been shown to have a bimodal distribution with peaks of the distribution occurring in steeply sloping areas and in flat areas (Gamache, 2004; Falorni et al., 2005; Reuter et al, 2007). Where voids exist in any of the source DEMs, they should be filled using auxiliary data to improve the significance of the fused DEM. Spatial interpolation methods are commonly employed to fill voids, e.g., kriging, spline, inversed distance weighted (Chen and Yue, 2010). More specialised void filling algorithms include the Delta surface fill (DSF) method (Grohman et al., 2006) and generative adversarial networks (Gavriil et al., 2019; Qiu et al., 2019). Differences between the DEM and the auxiliary data would need to be resolved before void filling proceeds. These differences can occur in areas such as the horizontal and vertical datum, the spatial resolution, production errors, first or second-order trends and the spatial distribution of errors (Fisher and Tate, 2006; Hutchinson, 1989; Kääb, 2005; Reuter et al., 2007). In their study, Tran et al. (2014) performed void filling/elimination of artefacts using the fill and feather method while Robinson et al. (2014) adopted the Delta surface fill method.

*4.2.2.2 Filtering of non-ground points*

Forests and urban structures obstruct the realisation of the bare ground topography by satellite sensors (see Figure 7). In forested areas, the tree heights can be removed to realise the bare-earth surface. This removal of tree heights can be achieved using empirical methods based on tree height estimation, by assuming a constant tree height or using ancillary data (Gallant et al., 2012; Yamazaki et al., 2017; Nwilo et al., 2017; Polidori and El Hage, 2020). Other approaches for removing non-ground objects include methods based on stochastic properties (Kraus and Pfeifer, 1998) and mathematical morphology (Vosselman and Maas, 2001). The treatment of source DEMs for tree offsets is an optional pre-processing step and only a few authors incorporated it in their workflow, e.g., Petrasova et al. (2017), Tran et al. (2014), Tian et al. (2018) and Pham et al. (2018). However, for many applications such as hydrological modelling, it is paramount to generate a bare-earth DEM (Schiewe, 2003; Buckley and Mitchell, 2004).

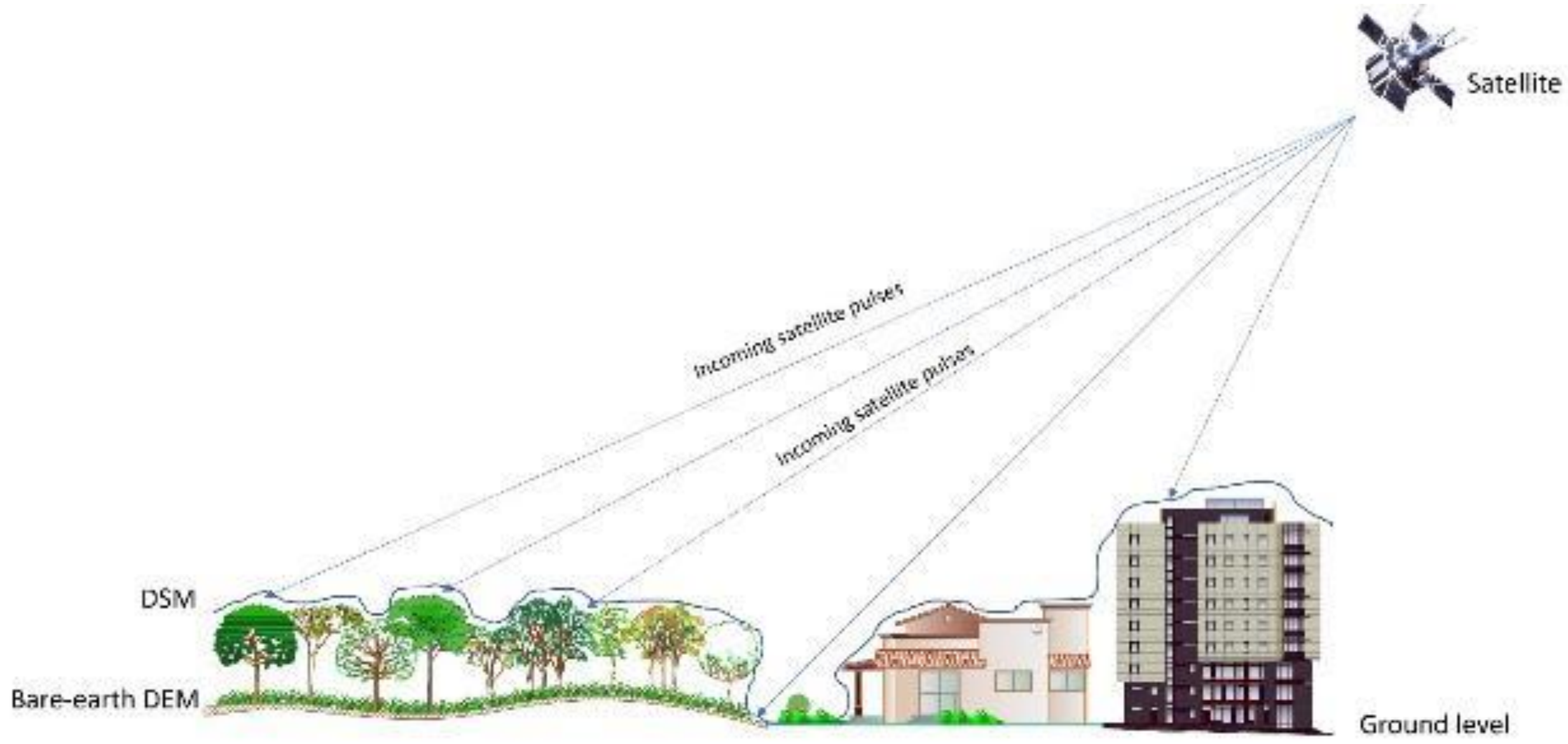

Figure 7: Illustration of the influence of above-ground offsets (buildings, trees and other non-ground points) which obscure the bare-earth topography in DEMs

*4.2.2.3 Vertical accuracy improvement*

DEM accuracy is influenced by several factors including the source data attributes, sensor distortions, land cover and terrain parameters (e.g., slope), errors inherent in the methods used for generating or producing the DEMs etc. (Li, 1992; Chen and Yue, 2010; Papasaika-Hanusch, 2012; Sadeq et al., 2016). There are techniques for improving the vertical accuracy of source DEMs, e.g., DEM difference surfaces (e.g., Xu et al., 2010) and outlier filtering. Also, by analysing the relationship between their vertical accuracies and land/topographic parameters (e.g., land use/ land cover, slope, aspect), the magnitude and distribution of the vertical error can be modelled and minimised. See for example, Wendi et al. (2016), Kulp and Strauss (2018) and Kim et al. (2020). Buckley and Mitchell (2004) have given a detailed explanation of errors that manifest in DEMs, including guidelines on their treatment.

*4.2.2.4 Spatial filtering/smoothing*

Filtering could involve the application of low-pass filters for removal of high-frequency artefacts (e.g., Karkee et al., 2008; Pham et al., 2018) or high-pass filters for removal of low-frequency artefacts (e.g., Karkee et al., 2008). Filtering approaches used in the investigated studies include the two-step mean filtering process (Bagheri et al., 2017; 2018) and adaptive multi-scale smoothing (Robinson et al., 2014).

### *4.2.3 DEM alignment*

#### *4.2.3.1 Resampling*

Input DEMs should be homogenised in terms of the pixel spacing/spatial resolution through the process of resampling. The usual approach is for the DEM with a coarser/larger grid to be resized to the same resolution as the DEM with a finer/smaller grid. However, additional errors (except random error) may be introduced when a larger grid DEM is resized to a smaller grid (Fisher and Tate, 2006; Wechsler, 2007; Fu and Tsay, 2016). This is taken into consideration in the work of Fu and Tsay (2016) in which a finer grid DEM generated from aerial LiDAR was adjusted to the same resolution as the coarser grid aero-photogrammetric DEM. The local topography and required applications for the fused DEM are factors that determine the common spatial resolution adopted (Petrasova et al., 2017). The grid size chosen has an impact on the DEM's accuracy and its ability to describe terrain features (Polidori and El Hage, 2020). The link between the DEM grid size and resolution is covered in the work of Polidori and El Hage (2020). Commonly used resampling (or interpolation) approaches in the reviewed studies include the bi-cubic convolution (Deng et al., 2019), regularised spline with tension (Petrasova et al., 2017) and bilinear resampling (Robinson et al., 2014).

#### *4.2.3.2 Harmonisation of spatial reference systems*

The projection or coordinate systems (CS) of source DEMs may vary (e.g., geographic CS or projected CS), and can be harmonised using projection tools in free/open-source (e.g., QGIS) or proprietary (e.g., ArcInfo) software. It may also be necessary to harmonise the DEMs to a uniform height system e.g., orthometric or ellipsoidal. To improve the stability of the DEMs, height normalisation could also be implemented (e.g., Bagheri et al., 2018).

#### *4.2.3.3 Co-registration*

The registration process is crucial in the fusion of DEMs with differing accuracy and spatial resolution (Ravanbakhsh and Fraser, 2013). Co-registration ensures that the source data coincide on a pixel-by-pixel basis and that they refer to the same ground location (Pohl and van Genderen, 2016). Registration involves a surface-to-surface matching in which vertical and horizontal offsets between the source DEMs are estimated and corrected (Ravanbakhsh and Fraser, 2013). It helps to compensate for translational and rotational discrepancies (Bagheri et al., 2018a). DEM co-registration approaches include manual registration, least-squares matching (Gruen and Akca, 2005) and iterative closest point registration. The iterative closest point algorithm for co-registration was adopted by Bagheri et al. (2018) and Deng et al. (2019).

#### 4.2.4 Regionalisation modelling

Since the best data sources are not always geomorphologically appropriate, regionalisation modelling is an applicable strategy for composing continuous elevation surfaces (that manifest natural and anthropogenic characteristics) and for modelling the quality of data sources (Podobnikar, 2005). Regionalisations can be obtained or modelled from several environmental variables such as land use/land cover, terrain attributes (slope, aspect, roughness, curvature etc.) and hydrology. These regionalisations can be linked to the pattern of height errors or residuals in the source DEMs and are valuable for preparing height error maps (HEMs) or predicting weight maps for fusion. According to Papasaika et al. (2008), slope, aspect and roughness are the most relevant attributes for geomorphologic analysis of DEMs.

Table 2 presents a summary of the pre-processing operations identified in the reviewed studies and frequency of use. The most applied pre-processing operations are resampling, co-registration, void filling, smoothing/filtering, and harmonisation of spatial reference systems.

Table 2: Pre-processing operations identified in the reviewed studies and frequency of use

| Pre-processing operation | No. of studies |
|---|---|
| Resampling/resizing (including interpolation) | 20 |
| Void/gap filling/masking or removal of invalid pixels | 8 |
| Smoothing/filtering | 7 |
| Co-registration (horizontal and/or vertical alignment) | 18 |
| Harmonisation of spatial reference systems (horizontal and vertical coordinate systems, datum transformation) | 12 |
| Height normalisation | 1 |
| Vertical alignment | 1 |
| Derivation of bare-earth DEM/removal or minimisation of elevation offsets | 4 |
| Error/blunder detection | 3 |
| SAR masking (layover and shadow, phase unwrapping error, regions of coherence lower than a certain threshold) | 3 |
| Water masking | 1 |
| Terrain classification | 1 |
| Slope and elevation thresholding | 1 |
| Minimisation of edge effects | 1 |
| Enhancement of terrain features | 1 |

### 4.3 Review of DEM fusion methods

Table 3 shows a list of DEM fusion methods and the relevant studies while Figure 8 shows the frequency of occurrence of the DEM fusion methods used in the studies. Weighted averaging is the most adopted method followed by variational methods. A detailed discussion of the methods is provided in the sections that follow.

Table 3: A list of DEM fusion methods and relevant studies. Some papers appear more than once due to their application of multiple methods

| **Fusion Method** | **Studies** |
|---|---|
| Simple averaging | Chu et al. (2017), Wang et al. (2018) |
| Weighted averaging | Ferretti et al. (1999), Gelautz et al. (2003), Stolle et al. (2005), Hoja et al. (2006), Papasaika et al. (2008), Xu et al. (2010), Schindler et al. (2011), Rossi et al. (2013), Robinson et al. (2014), Jain et al. (2014), Tran et al. (2014), Karakasis et al. (2014), Deo et al. (2015), Gruber et al. (2016), Fu and Tsay (2016), Petrasova et al. (2017), Bagheri et al. (2017b), Bagheri et al. (2018a), Bagheri et al. (2018c), Deng et al. (2019), Ajibola et al. (2019), Du et al. (2019), Arief et al. (2020) |
| Adaptive regularisation variation model based on sparse representation | Guan et al. (2020) |
| Bayesian inference | Sadeq et al. (2016) |
| Maximum likelihood | Jiang et al. (2014), Sadeq et al. (2016) |
| Clustering approaches | Fuss et al. (2016) |
| Coherence weighted average, Maximum coherence, Coherence mixture decision | Chaabane (2008) |
| Combination of maximum rule and maximum correlation rule | Gamba et al. (2003) |
| Combination of preferential rule and weighted fusion | Deng et al. (2011) |
| Convex energy functional (with β-Lipschitz continuous gradient) | Perko and Zach (2016) |
| Fast Fourier Transform (FFT) | Karkee et al. (2008) |
| Guided filter | Dong et al. (2018) |
| Kalman filter | Slatton et al. (2001), Zhang et al. (2016) |
| Linear combination approach for weight estimation | Pham et al. (2018) |
| Mathematical approach using weights | Papasaika et al. (2009) |
| Multiple-point geostatistical simulation | Tang et al. (2015) |
| Multi-scale decomposition and a slope | Tian et al. (2018) |
| position-based linear regression | |
| Regularised super-resolution (SR) | Yue et al. (2015) |
| Self-consistency measures | Schultz et al. (1999), Stolle et al. (2005) |
| Sparse representations supported by weight maps | Papasaika et al. (2011), Schindler et al. (2011) |
| Variational models | Kuschk et al. (2017), Bagheri et al. (2017b), Bagheri et al. (2018c) |
| Spectral methods | Karakasis et al. (2014) |

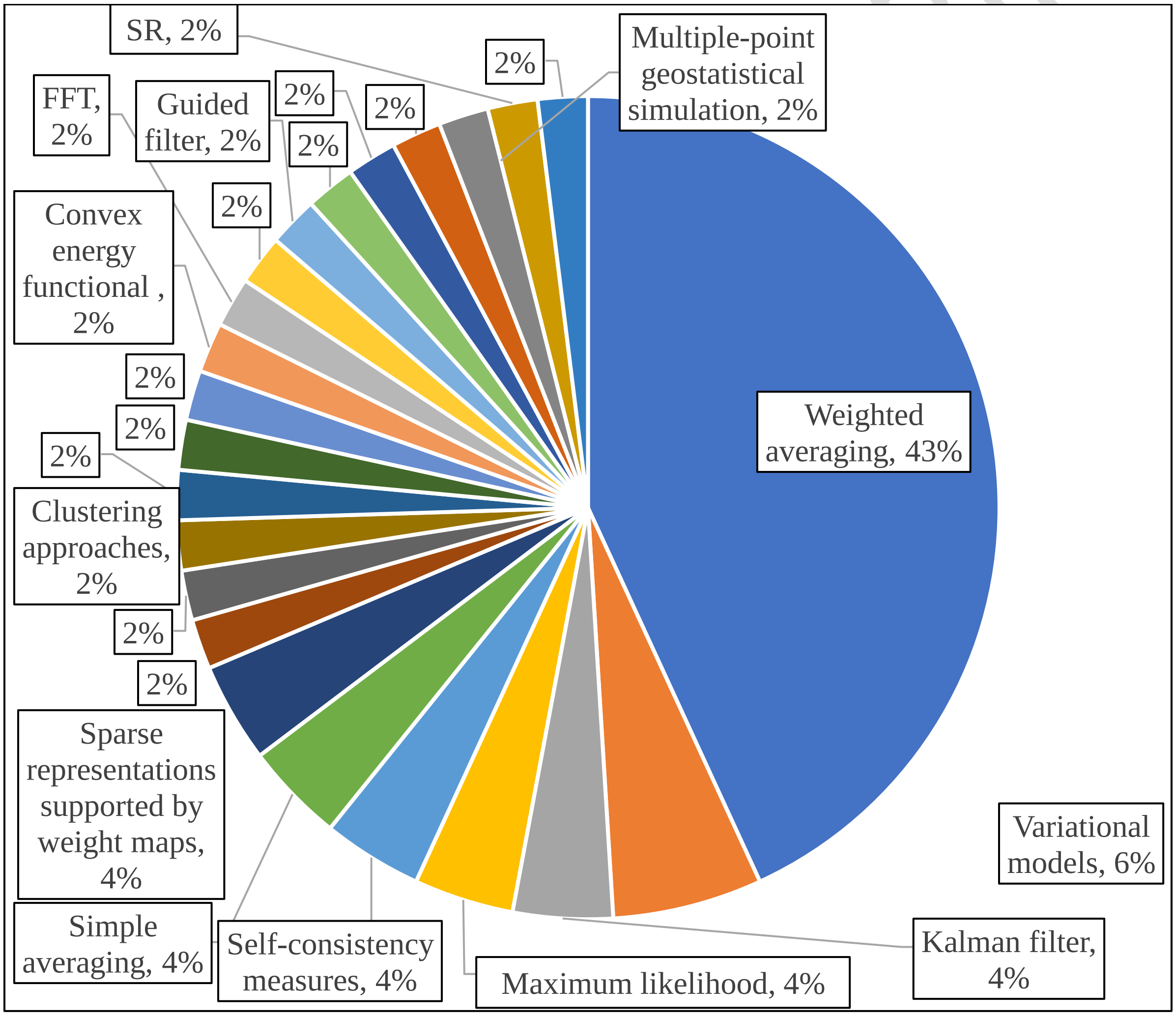

Figure 8: Frequency of occurrence of the DEM fusion methods in the reviewed studies. Due to its simplicity and ease of implementation, weighted averaging is the most frequently adopted method.

### *4.3.1 Simple averaging*

Simple averaging consists of calculating the mean elevations of the input DEMs on a cell-by-cell basis (Leitão and Sousa, 2018). It works well if the images are acquired from the same sensor (Banu 2011; Kaur et al., 2021). Wang et al. (2018) applied the simple averaging method in the fusion of ASTER GDEM and elevations from the Ice, Cloud, and land Elevation (ICESat) sensor in Mertz Glacier, East Antarctica. The simple averaging was applied at pixels where elevations from both DEMs coincided, and with absolute height difference ≤ 10m. Surfaces outside these regions were filled up by ICESat-only or ASTER-only heights depending on whichever was available. A final DEM was then generated from all available height points. Similarly, Chu et al. (2017) employed a simple averaging procedure to fuse spotlight conventional DEMs, and small baseline subset (SBAS) DEMs generated from Radarsat-2 using equation 1:

$$h(x) = \frac{\sum_i^n h_{i(x)}}{n} \qquad (1)$$

Where $h(x)$ is the mean topographic height at the pixel location, $h_i(x)$ is the topographic height of i'th DEM, and n is the number of DEMs.

They reported considerable improvements and high-quality DEMs for their study area, with a root mean square error (RMSE) of 9.7 m through the fusion of the spotlight and SBAS DEMs.

### *4.3.2 Weighted averaging*

An optimal solution is not always achieved by averaging all the corresponding elevation points to derive new heights (Hoja and d'Angelo, 2009; Tian et al., 2018). In weighted averaging (WA) fusion, weights are used to quantify the influence of the input DEMs at every grid or surface location and are a function of the relative height accuracy (Schindler et al., 2011). Weighted averaging is a commonly used approach. Several researchers have established that it is a simple approach for DEM fusion with low computational cost (e.g., Hoja and d'Angelo, 2009; Gruber et al., 2016; Deo et al., 2015; Bagheri et al., 2018). Its performance is dependent on the weights that describe the distribution of height error within the pixels of the DEMs being fused (Roth et al., 2002; Bagheri et al., 2018b).

In practice, weighted averaging has been shown to give acceptable results (Tran et al., 2014; Gruber et al., 2016; Fu and Tsay, 2016; Bagheri et al., 2018a; Deng et al., 2019). The variation in the weights is influenced by several factors such as scene characteristics, sensor technology and the methods used for generating the source DEMs (Schindler et al., 2011). However, even slight elevation differences (in the order of a few metres) between datasets can result in rough edges at the fused interfaces (Deng et al., 2019). In WA fusion, the weights are usually estimated based on the error between the DEMs being investigated and the reference data. However, this can be challenging since reference data sources (and height error maps) are not always readily available (Papasaika et al., 2011; Pham et al., 2018). Bagheri et al. (2018a) has also pointed out that WA is not an optimal solution for areas with complex terrain morphology such as urban areas due to the presence of high-frequency contents e.g., building edges. This could lead to poor representation of the sharp outlines of buildings, along with the transfer of noise effects which spoil the characterisation of the buildings in the fused DEM. According to Schindler et al. (2011), "the correct choice of weights has a far greater influence on the results than the mathematical recipe for fusing the inputs".

Generally, the weighted mean of a non-empty set of data $fx_1$; $fx_2$; … $x_n$ with non-negative weights $fw_1$; $fw_2$; :::; $w_n$, is given by Papasaika-Hanusch (2012):

$$\bar{x} = \frac{\sum_{i=1}^{n} w_i x_i}{\sum_{i=1}^{n} w_i} = \frac{w_1 x_1 + w_2 x_2 + \cdots + w_n x_n}{w_1 + w_2 + \cdots + w_n} \qquad (2)$$

Points or data elements with a higher weight will have a larger influence on the weighted mean. The formula can be simplified by normalising the weights such that they sum up to 1 (i.e., $\sum_{i=1}^{n} w_i = 1$) (Papasaika-Hanusch, 2012).

A few key points on WA fusion are summarised below:

1. It deals with each grid independently without consideration for the resulting surface shape (Schindler et al., 2011). However, this weakness was partly overcome in the work of Fu and Tsay (2016) where the precision was not determined at every grid point, but weights were assigned to grids based on their class of terrain surface. This weakness of WA fusion was also mitigated in the work of Papasaika et al. (2011) with the use of sparse representations.
2. There is the potential for artefacts to be introduced in the fused DEM through up-sampling from one of the input DEMs (Schindler et al., 2011).

3. The precise determination of the fusion weights is more important than the choice of the mathematical toolbox to be used in the fusion (Schindler et al., 2011).

Due to its simple implementation, WA is frequently adopted for the fusion of global DEMs. Jain et al. (2014) adopted WA fusion based on height error maps for merging ascending/descending pairs of TanDEM-X SAR co-registered Single Look Slant Range Complex (CoSSC) at sites in Mumbai, western coast of India. After fusion, the RMSE reduced by 2 - 4 m and there was an 84% reduction in the number of voids.

Deo et al. (2015) applied a mathematical WA approach using TanDEM-X acquisition geometry in ascending and descending pass to optimize the incidence angle for selection of input TanDEM-X pairs in Mumbai, Maharashtra State, India. The fusion proceeded based on an optimum weight selection approach that used a Height Error Map (HEM) and a robust layover shadow mask. After fusion, there was a significant reduction in the number of invalid pixels (from 5.02% and 6.34% in the ascending and descending pass DEMs respectively to 2.14% in the fused DEM). The RMSE decreased from 7.3m in the ascending DEM and 7.5m in descending DEM to 7.0m in the fused DEM. In the fused DEM, 90% relative error value was reduced by 6% and 90% absolute error value decreased by 4%. There was also a slight improvement in the standard deviation of the error in the fused DEM by about 5%. Bagheri and colleagues (Bagheri et al., 2017b, 2018a, 2018c) implemented weighted averaging in several fusion experiments. For example, Bagheri et al. (2018a) applied machine learning techniques to WA fusion by employing sophisticated ANN-predicted weight maps to fuse TanDEM-X and Cartosat-1 DEMs in Munich, Bavaria. Training data was limited to land types within urban areas and surroundings. They achieved up to 50% improvement in the relative accuracy of DEMs in urban areas, and 22% in non-urban areas. There was an improvement in the quality of fused DEMs based on the ANN-predicted weights.

Gruber et al. (2016) implemented an improved WA fusion approach based on priority values of the most reliable InSAR DEM heights at different sites in the United States, Russia and Switzerland. Height inconsistencies in input DEM acquisitions were detected leading to improved vertical accuracies of the mosaic. The occurrence of unreliable height values was significantly reduced compared to the common WA method for TanDEM-X mosaics. They implemented the fusion concept on various test areas affected by phase unwrapping (PU) errors in both flat and mountainous terrain, as well in forested areas affected by height discrepancies, and the results showed improvements in the quality of the fused TanDEM-X DEM. Deng et al. (2019) used an adaptive WA algorithm with multiple search distances for fusing TanDEM-X, Terrestrial Radar Interferometry (TRI) and Structure-from-motion (SfM)-derived DEMs over the Nevado del Ruiz volcano in Colombia. Significant improvements in the representation of terrain details were achieved in the fused DEM, including smooth elevation transitions at the edges of input datasets.

Robinson et al. (2014) applied gaussian WA fusion along a very wide area in the N59 - N60 latitudinal band in the development of the enhanced global DEM, EarthEnv-90m. The weight was a function of distance from the transition line between two source DEMs - ASTER GDEM2 and SRTM v4.1. The spatial resolution of the fused DEM (90m) was constrained by the resolution and quality of the input datasets. Elevations in EarthEnv-DEM90 were on average, 0.82m higher than the control point measurements while ASTER GDEM2 elevations were on average, 3.08 m lower. Petrasova et al. (2017) proposed a generalised fusion approach in which the transition between DEM overlap zones was controlled by distance-based WA with spatially variable width based on elevation differences. They applied it to an SfM-derived DEM and a LiDAR DEM from the North Carolina Floodplain Mapping Program. In the fused DEM, there was a seamless transition in the overlap zone of the input DEMs. The use of a data-driven spatially variable overlap width achieved a better transition between the two merged DEMs, improved the preservation of terrain shape and micro-topography, and minimised edge artefacts. However, their approach did not derive a 'better' terrain representation globally; the fusion was only applied locally along the edges of the DEMs.

Tran et al. (2014) applied WA for fusion of 30m ASTER GDEM v2 and 90m SRTM v4.1 based on a landform classification map. The fusion achieved an RMSE reduction from 14.9m in ASTER GDEM and 14.8m in SRTM to 11.6m in the fused DEM, and the terrain-related parameters from the fused DEM were comparable to reference data. Gelautz et al. (2003) merged stereoscopic and interferometric DEMs from Radarsat-2 and ERS-2 by applying a user-defined weighting function to a filtered coherence map. In the fusion, the most severe interferometric errors were successfully replaced by more robust stereo measurements. The fused DEM displayed a significantly lower error rate than the individual DEMs. In Ferretti et al. (1999), ERS-1/2 tandem data were fused by means of WA carried out in the wavelet domain. This yielded a more reliable fused DEM with reduced elevation error dispersion since the uncorrelated atmospheric and noise phase contributions from the individual interferograms were averaged.

Other studies on WA fusion include Du et al. (2019) – fusion of TanDEM-X/TerraSAR-X pairs; Arief et al. (2020) - fusion of ascending and descending pass TanDEM-X/TerraSAR-X in stripmap mode; Ajibola et al. (2019) - fusion of SfM-derived DEMs from rotorcraft and fixed-wing unmanned aerial vehicles (UAVs); Rossi et al. (2013) - TanDEM-X and Cartosat-1 fusion; Schindler et al (2011) - ALOS-SPOT, ERS-SPOT and ALOS-ERS fusion; Xu et al. (2010) - fusion of TopoSAR and ALOS PRISM; Papasaika et al. (2008) - fusion of IKONOS and airborne LiDAR DEMs; Hoja et al. (2006) - fusion of SPOT HRS stereo data, C-band and X-band SRTM, ERS data; Stolle et al. (2005) - fusion of IKONOS DEMs in which the weights are the inverse of self-consistency differences; Fu & Tsay (2016) - fusion of aero-photogrammetric DEM and airborne LiDAR DEM; and Karakasis et al. (2014) - weighted average of expansion coefficients of source DEMs.

### *4.3.3 Variational models*

Variational models have found wide applications in signal and image denoising (Rudin et al., 1992; Nikolova, 2004; Wang et al., 2020), image fusion (Zhou et al., 2011; Ballester et al., 2006; Liu and Ding, 2019), image reconstruction (Williams et al., 2016), image segmentation (Zhai et al., 2010; Liu et al., 2018), image enhancement (Zhou et al., 2010; Williams et al., 2016; Wang et al., 2020), and image classification and restoration (Samson et al., 2000). An introduction to variational image-processing models and applications is given by Chen (2013), and Vese and Le Guyader (2015) have also written a textbook on variational methods in image processing. The total variation-based model (TV) is a common type of variational model in which the gradient of an output image is chosen to form the regularisation term based on various norms (Bagheri et al., 2018). The TV-based variational model has several advantages. For example, its convexity guarantees the finding of a solution through minimisation of the energy functional, it is well suited for parallelisation, and with recent advances in high-performance computing, can lead to efficient algorithms (Kuschk and d'Angelo, 2013; Bagheri et al., 2018).

The general idea of TV-based regularisation which was proposed by Rudin et al. (1992) was later combined with a robust data term labelled by the $L_1$ norm (TV-$L_1$) (Pock et al., 2011) (Eq. 3). The left term in the formula is the regularisation term and it guarantees a smooth solution and the $L_1$ data term constrains the solution to conform to each of the input DEMs. The two terms of the energy functional are balanced by the trade-off parameter ($\alpha$). In the energy functional, $d_i$ refers to the input DEMs while $u_f$ is the unknown fused DEM to be solved in the minimisation process of the energy functional.

$$min_{uf}\{ \alpha \,||\, \nabla u_f \,||_1 + \sum_{i=1}^{m} ||\, u_f - d_i \,||_1 \} \quad (3)$$

A higher-order extension of total variation known as Total Generalised Variation (TGV) was later introduced by Bredies et al. (2010). TGV constrains the solution to consist of piecewise planar functions and unlike the TV-$L_1$ model, it includes slanted planes (Pock et al., 2011; Kuschk and d'Angelo, 2013).

Although the TV-$L_1$ is robust against outliers and preserves edges, it suffers from the staircasing effect which manifests as artificial discontinuities in the output DEM, especially in high-resolution DEMs (Pock et al., 2011; Chan et al., 2006; Bagheri et al., 2018). Coupled with this, the $L_1$ is not always the optimal choice in all data fusion scenarios. Another variational model (the Huber model) has thus been proposed as an alternative to correct the drawbacks of the TV-$L_1$ model (Pock et al., 2011; Bagheri et al., 2018). The Huber model applies to the Huber norm in both fidelity and penalty terms (Huber, 2011; Bagheri et al., 2018).

Kuschk and colleagues (Kuschk and d'Angelo, 2013; Kuschk et al., 2017) and Bagheri and colleagues (Bagheri et al., 2017, 2018) have taken advantage of the TV regularisation for DEM fusion. Kuschk and d'Angelo (2013) specifically adapted a version of TV-$L_1$ that incorporates appropriate weight maps in the fusion algorithm (Eq. 4). They applied a weighted version of TV-$L_1$ and TGV-$L_1$ for the fusion of Worldview 2 stereo DEMs in London and CartoSat-1/Worldview-2 derived DEMs at three ISPRS Benchmark sites in Spain (La Mola, Vacarisses and Terassa). In the evaluation, significant improvements were achieved in the fused DEM. In related work, Kuschk et al. (2017) achieved a robust DEM fusion by incorporating explicit surface priors and local planarity assumptions and by minimising the $L_1$-distance of the fused DEM pixels to that of the input DEMs

$$min_{uf}\{ \alpha \,||\, \nabla u_f \,||_1 + \sum_{i=1}^{m} w_i ||\, u_f - d_i \,||_1 \} \quad (4)$$

Bagheri et al. (2017b) implemented TV-$L_1$ fusion on CartoSat-1 and TanDEM-X DEMs in Germany, based on weight maps derived by a specifically trained artificial neural network (ANN). They evaluated the accuracy of the fused DEM using groundtruth LiDAR and observed that TV-

$L_1$ achieved a higher improvement compared to the weighted averaging method. The quality of the fused DEM was further enhanced with the weighted version of TV-$L_1$. In another experiment, Bagheri et al. (2018c) employed fusion based on the TV-$L_1$ model and the Huber model. Their results illustrated the superiority of variational models over the classic WA in terms of reduction of noise effects and other quality considerations.

#### *4.3.4 Super-resolution (SR) methods*

Image Super resolution (SR) can generate an image with higher resolution by exploiting the redundancy of information contained in multiple images of low-resolution (Park et al. 2003; Yue et al., 2015). It considers the limited resolution and degradation factors in the source DEMs to produce a fused DEM with higher resolution thus achieving a trade-off between spatial coverage and resolution. SR has been applied in image reconstruction with variable resolution (Joshi et al., 2005; Song et al., 2015) and has other potentials. A review of image super-resolution techniques and applications is given by Yue et al. (2016) while Liu et al. (2021) reviewed the development of traditional, learning-based, and deep-learning-based methods of SR reconstruction of remote sensing images.

Inspired by SR, Yue et al. (2015) implemented a regularised SR framework in different fusion experiments using the 90m SRTM v4.1, 30m ASTER GDEM2, 12m WorldDEM (a product of the TanDEM-X Mission) and a 10m C-band TOPSAR DEM in Australia and USA. Their method combined super resolution, void filling, and noise suppression into a universal framework. The resulting DEMs were reconstructed using the low-resolution data as fundamental information and part of the high-resolution data as a constraint. Several attributes were observed in the fused DEMs. For example, the removal of noise and seamless fusion of the high- and low-resolution information (WorldDEM - ASTER fusion), more comprehensive and accurate terrain details (Airborne InSAR – ASTER fusion), and a more accurate bank line including visually attractive contour lines, continuous terrain, and comprehensive terrain features (Airborne InSAR – ASTER - SRTM fusion).

#### *4.3.5 Kalman filter*

The Kalman filter (named after Rudolf E. Kalman) is one of the most common data and sensor fusion algorithms (Alsadik, 2019). It is an efficient optimal estimator with a recursive computational method that can determine the state of a discrete-data controlled process from noisy measurements, and providing in addition, an estimation of the uncertainty of the estimates (Thomson and Emery, 2014). The Kalman filter was originally designed for monitoring real-time processes with the goal of filtering noisy observations (Kalman, 1960; Roujean, 2018). Chou et al. (1994) introduced a framework for modelling stochastic phenomena at variable scales based on a Multi-Scale Kalman filter (MKS). The MKS was subsequently applied by Slatton et al. (2001) for the fusion of variable resolution NASA/JPL Topographic SAR (TOPSAR) and LiDAR data from the Optech Airborne Laser Terrain Mapping (ALTM) instrument in the Bolivar Peninsula of Texas, USA. The fusion achieved an improvement in the accuracy with a global mean square error (MSE) reduction of 88-90% for bare surface heights and 66-87% for vegetation heights.

Relatedly, Slatton et al. (2002) applied the MKS framework to fuse DEMs from Tandem ERS data, a single TOPSAR flight line from the PacRim mission and from three TOPSAR flight lines which were acquired as part of the PacRim 2000 TOPSAR deployment. The height uncertainty of the fused DEM was considerably lower than that of the individual InSAR acquisitions, and there was a monotonic decrease in the mean uncertainty. A dramatic improvement was also observed in steep channels where shadowing and foreshortening could be problematic. In more recent work, Zhang et al. (2016) has used an Extended Kalman Filter (EKF) for the fusion of Multi-baseline and Multi-frequency Interferometric Results (MMIRs) derived from TerraSAR-X and Envisat ASAR data. To generate the fused DEM, the InSAR DEMs were taken as states in the prediction step and the flattened interferograms were taken as observations in the control step. The fused DEM was much more accurate than the input DEMs but not without deficiencies caused by layover on steep front slopes.

#### *4.3.6 Multiple-point geostatistical simulation*

A multiple-point geostatistical simulation (MPS) has been proposed as an advancement of traditional geostatistics (Guardiano and Srivastava, 1993; Rasera et al., 2015). Instead of relying on variograms for quantifying the spatial structure related to pairs of locations, the MPS adopts structures from the training images, from which local patterns of the site can be captured (Tang et al., 2015). The original implementation of MPS builds on the paradigm of the traditional indicator simulation (Deutsch and Journel, 1998; Tang et al., 2015). The main applications of geostatistical simulation methods for topographic modelling are in the aspects of downscaling and assessment of errors in DEMs (Zakeri and Mariethoz, 2021). For example, Rasera et al. (2019) proposed an MPS approach for downscaling coarse Swisstopo ALTI3D DEM. Applications in DEM error assessment include the works of Chu et al. (2014) and Leon et al. (2014). Furthermore, MPS-based methods have been utilised for image fusion and super-resolution enhancement (e.g., Tang et al., 2015b). MPS can establish non-linear relationships from various data sources and can thus provide a solution to the problem of DEM fusion by incorporating multiple-point correlation (Tang et al., 2015a).

The application of MPS-based models to DEM fusion was demonstrated by Tang et al. (2015a). The fusion performed in their work was for data refinement i.e., spatial downscaling of coarse-resolution Global Multi-resolution Terrain Elevation Data (GMTED) through the use of co-registered, and sparsely sampled fine-resolution SRTM data. They proposed using FILTERSIM for the fusion of coarse-resolution GMTED2010 from China with fine-resolution SRTM data. FILTERSIM uses a filter-based prototyping process for deriving the fused DEM. Six filters in a local window are used: the average, gradient and curvature along both the east-west and north-south directions. Since the conventional FILTERSIM does not consider the non-stationarity of spatial structures when gathering similar patterns in the prototyping process, they developed a feature-based prototyping process wherein new filters were incorporated into FILTERSIM. In the modified FILTERSIM algorithm, instead of a filter-based prototyping process, terrain features are derived using the: (i) DEM residual surface, (ii) vector ruggedness measure; and (iii) ridge valley class. The fused DEM generated with the improved FILTERSIM algorithm achieved greater accuracy and led to better preservation of spatial structures.

#### *4.3.7 Multi-scale decomposition*

Multi-scale decomposition is an important analytical approach with applications in data fusion. Two multi-scale decomposition methods – the wavelet transform (WT) and the empirical mode decomposition (EMD) are commonly used in image fusion. WT is a multi-resolution time-frequency analysis approach with high time resolution in the high-frequency section and high frequency resolution in the low-frequency section (Aghajani et al., 2016; Belayneh et al., 2016; Tian et al., 2018). The WT has auto-focusing ability, is adaptable and well suited for non-stationary signal processing (Tian et al., 2018). The expressions for the WT can be found in Mallat (2009) and Daubechies et al. (1998). EMD was proposed by Huang et al. (1998). It is a data-driven technique that preserves the characteristics of the signal, and has applications in smooth processing, and in non-linear, non-stationary signal linearisation (Tian et al., 2018). The Bidimensional Empirical Mode Decomposition (BEMD) is an extension of EMD that is used for 2D image analysis and processing (e.g., Nunes et al., 2003). More details on image decomposition using BEMD are given in Nunes et al. (2003) and Equis and Jacquot (2009). Another image decomposition method known as nonlinear adaptive multi-scale decomposition (N-AMD) was initially developed to remove the cycle from sunspots but has shown superiority over existing methods (e.g., wavelet-based and chaos-based techniques) in trend determination and noise removal (Gao et al., 2010; Hu et al., 2009; Tung et al., 2011; Tian et al., 2018). N-AMD is effective at reducing the noise of time-series data and can determine any trend signal without prior knowledge (Tung et al., 2011; Tian et al., 2018).

Tian et al. (2018) applied four multi-scale decomposition methods (Bior3.7 wavelet, Haar wavelet, BEMD and N-AMD) to conduct a rule-based fusion of the 30m SRTM1 and 30m AW3D30 DEMs at two sites in southwestern and northern China. The fused DEM was further enhanced with the application of a slope position-based linear regression. In the fused DEMs, texture and details were clear and there was no mutation region. N-AMD decomposition method which showed the best performance, can effectively detect the noise of two DEMs and ensure the consistency of trend signals. Figures 9 and 10 show the fused DEMs obtained using the Bior3.7 wavelet, Haar wavelet, BEMD, and N-AMD methods.

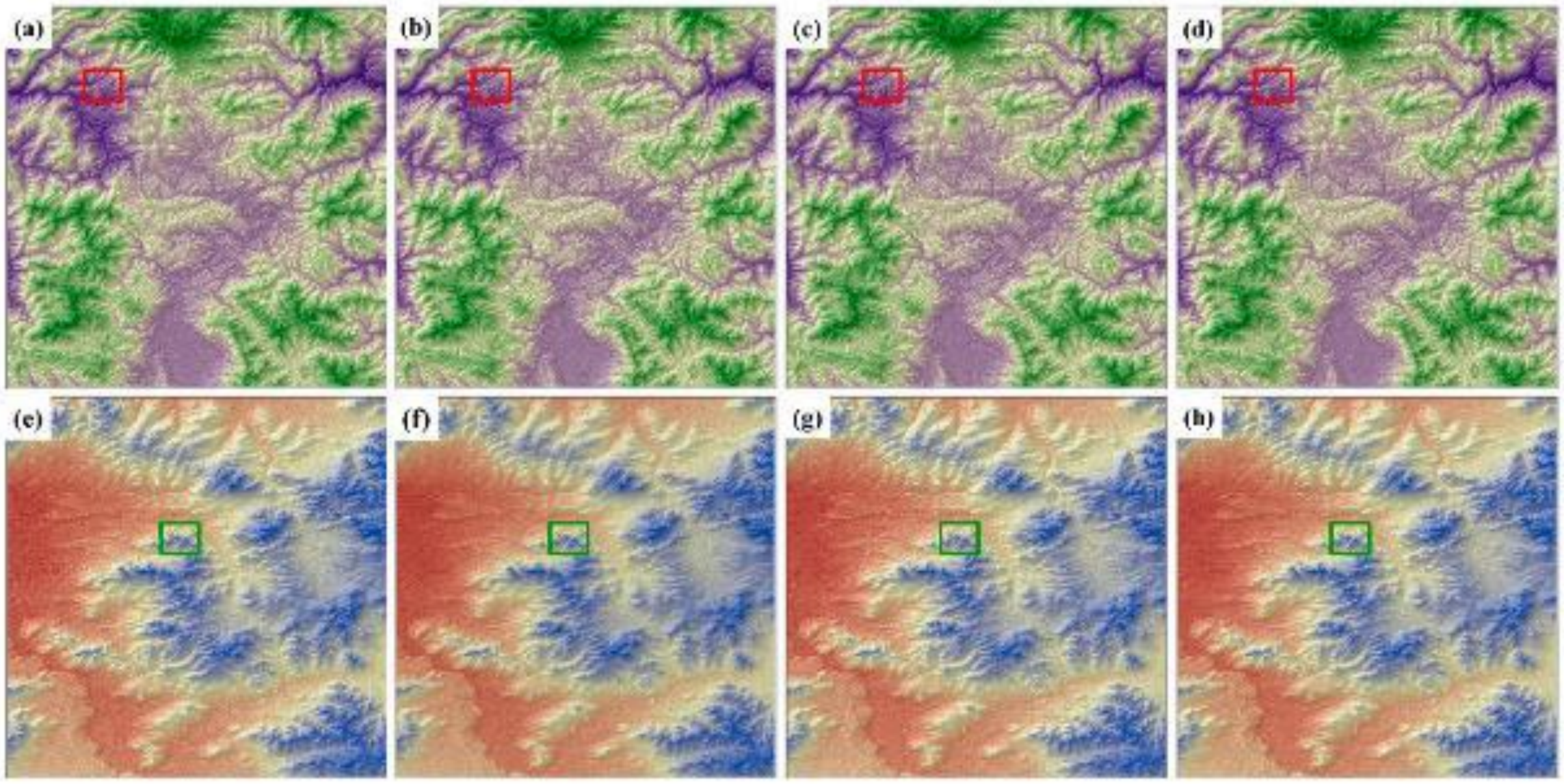

Figure 9: The fused DEMs obtained using the Bior3.7 wavelet, Haar wavelet, BEMD, and N-AMD methods at site A (a–d) and site B (e–h). The red and green rectangles are the areas selected in sites A and B that are enlarged in Figure 10. (Source: Tian et al., 2018)

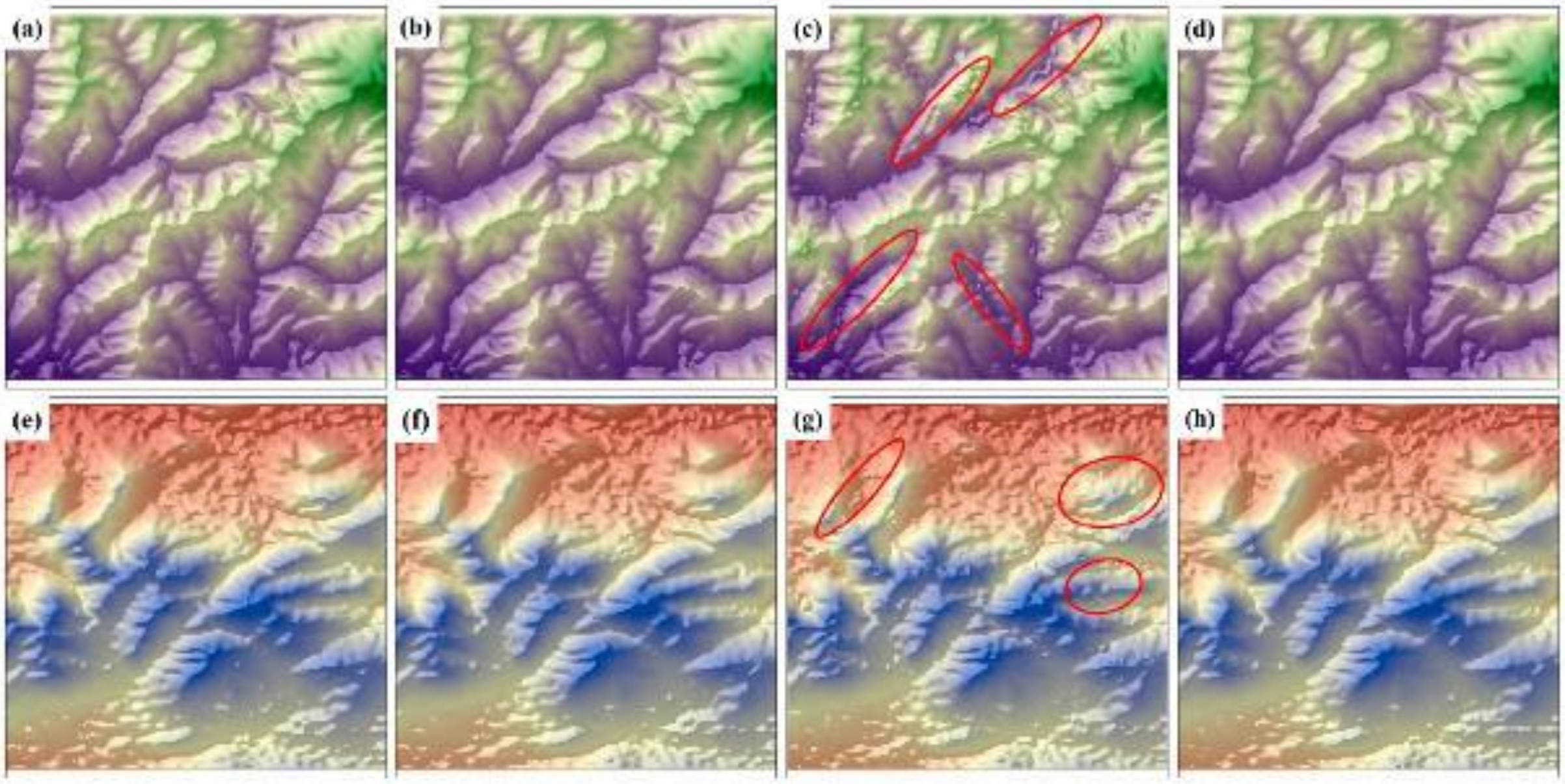

Figure 10: Enlarged images of the fused DEMs obtained using the Bior3.7 wavelet, Haar wavelet, BEMD, and N-AMD methods at site A (a–d) and site B (e–h). The red shapes indicate irregular patches. (Source: Tian et al., 2018)

### *4.3.8 Other methods*

When the sample data is limited, it may be preferable to employ Bayesian statistics and exploit *a priori* values in a calculation to minimise the uncertainty (Berry, 1997; FDA, 2010; Sadeq et al., 2016). Researchers have successfully applied Bayesian approaches in remote sensing for image fusion, image segmentation and image classification (e.g., Mascarenhas et al., 1992; Shi and Manduchi, 2003; Zhang, 2010). Bayesian approaches can reduce the effect of noise in datasets by using suitable *a priori* knowledge (Sadeq et al., 2016). Only Sadeq et al. (2016) was found to have adopted a Bayesian approach for fusing DEMs. In their work, they fused 1-metre DEMs from Worldview-1 and Pleaides covering parts of Glasgow in the United Kingdom and compared the results with a maximum likelihood fusion using the same datasets. The maximum likelihood method maximizes the probability associated with the estimated value of a pixel in the fused DEM. Their findings showed that the Bayesian approach had more influence on the resulting DEM than maximum likelihood. Another observation was that the Bayesian approach helped to smoothen the surface of the generated structures. However, the errors (discrepancies) between the fused DEM and Global Navigation Satellite System (GNSS) checkpoints were lower for the maximum likelihood than the Bayesian approach. Another application of maximum likelihood was in the work of Jiang et al. (2014) where the variance of phase noise and the height of ambiguity were used for optimally determining the fusion weights.

The concept of DEM fusion in the frequency domain was tested by Karkee et al. (2008) on 90m SRTM v4.1 and 30m ASTER GDEM. After co-registration and void-filling, the DEMs were converted to the frequency domain and low-pass and high-pass filters were applied to filter out errors in the frequency ranges. The filtered DEM spectra were summed in the frequency domain and then converted to the spatial domain. They observed a 42% improvement in the RMSE of the fused DEM. There was an RMSE reduction to 16.4m from 18.3m in SRTM and 28.3m in ASTER (i.e., 42% reduction in error with respect to ASTER relative DEM and 10% improvement in RMSE of SRTM). The fused DEM showed improvements in accuracy and completeness over the individual DEMs. Figure 11 shows the result of the fusion process.

The concept of sparse representation has previously been applied to super-resolution problems (Yang et al., 2010), and image denoising (Elad and Aharon, 2006) with satisfactory performance. This concept was exploited by Papasaika et al. (2011) and Schindler et al. (2011) to improve the quality of DEMs through the framework of sparse representations supported by weight maps. According to Guan et al (2020), “the DEM reconstruction by sparse representation assumes that the topography of the DEM can be represented by image atoms with similar structural features. This process includes dictionary training and sparse representation”.

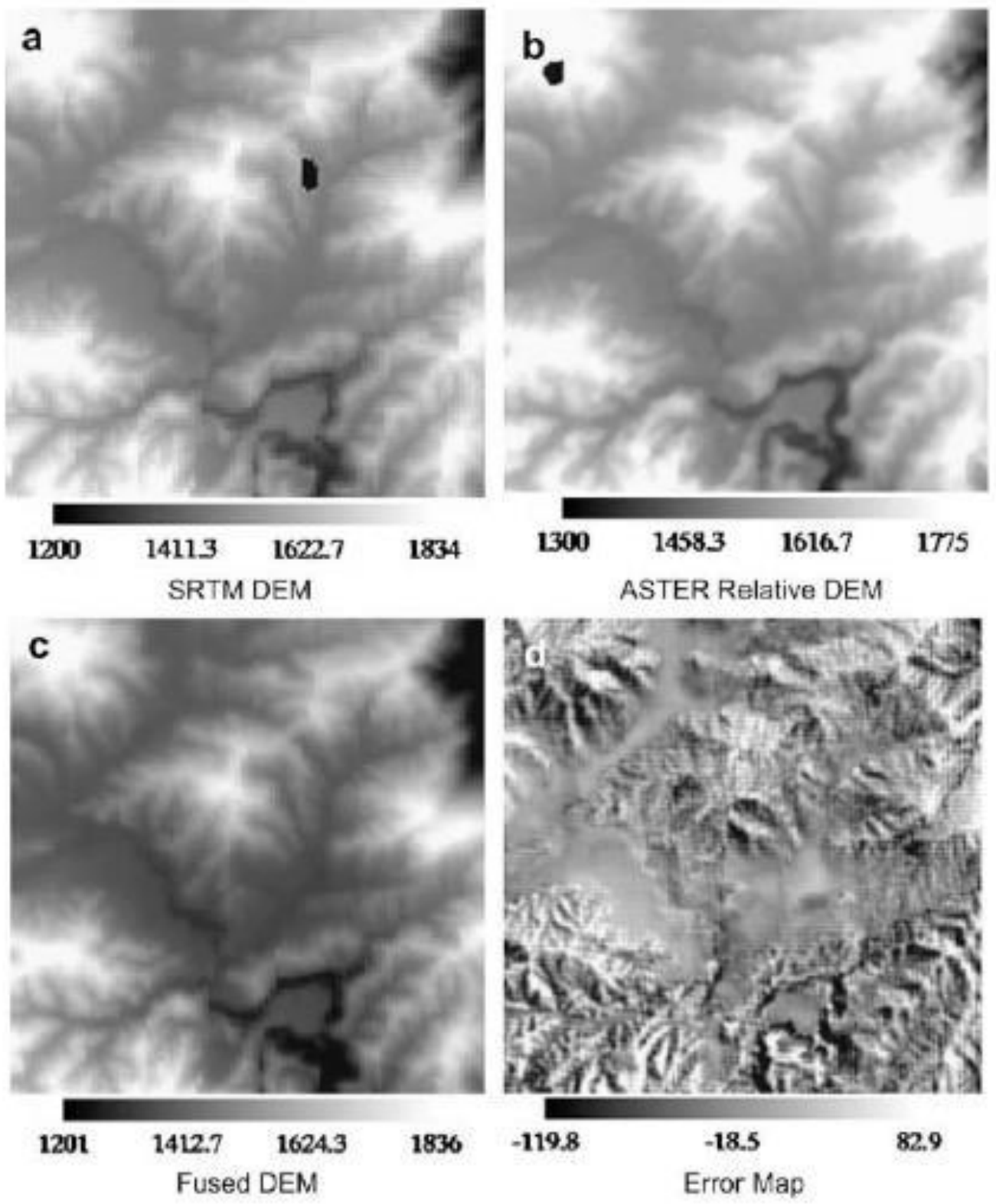

Figure 11: Result of the fusion process by Karkee et al. (2008) (elevation in metres). (a) SRTM DEM, (b) ASTER relative DEM co-registered with SRTM coordinates and histogram shifted to the mean elevation of SRTM, (c) fused DEM, and (d) error map. The darker regions in (d) represent the higher negative errors and vice versa. The gray region was the area with lower absolute error.

The concept of self-consistency measures to identify unreliable points in a distribution was first introduced by Leclerc et al. (1998a) and applied for the detection of terrain changes in Leclerc et al. (1998b). The idea was extended by Schultz et al. (1999) to detect unreliable elevation points in DEMs generated from stereoscopic image pairs and to subsequently fuse the DEMs. Self-consistency is based on two key factors: (i) unreliable elevation estimates (outliers) are first identified using the concept of self-consistency and excluded from further processing, (ii) fusion is then performed on the reliable points to produce an optimal elevation model. In Schultz et al. (1999), a significant reduction of residual errors was achieved in the fused DEM through self-consistency, and no apparent anomalies were detected. This method was also used in Schultz et al. (2002) and Papasaika et al. (2011a,b).

Fuss et al. (2016) were the first to use a k-means clustering algorithm for DEM fusion. They applied it to multiple, overlapping Radarsat-2 fine-quad mode, single look complex (SLC) imageries, with varying incidence angles, extracted by same side stereo-radargrammetry. The k-means approach exploited the consistency in the estimates as indicators of precision and accuracy. Their fusion approach incorporated slope and elevation thresholding, k-means clustering of the elevation estimates at each pixel location, as well as filtering and smoothing of the fused DEM. They thereafter compared their method with fusion by simple averaging. The standard deviation (SD), mean absolute error (MAE) and RMSE reduced from 5.9m, 6.6m and 8.0m in the DEM fused by simple averaging to 5.2m, 5.3m and 6.4m respectively in the final DEM fused by the k-means clustering approach.

In Chaabane (2008), three conjunctive operators were applied for DEM fusion namely: coherence weighted average (fusion by a simple coherence weighted averaging operator), maximum coherence (fusion based on a conjunctive operator which opts for pixels with maximum coherence), and coherence mixture decision (fusion based on coherence information differences). In Gamba et al. (2003), a combination of maximum rule and maximum correlation rule was used for DEM fusion. This is a mixed approach that combines height values from different DEMs using two rules: (i) maximum rule - considers only the maximum height from different DEMs, and (ii) maximum correlation rule - selects the highest height value, provided the corresponding DEMs also have the highest correlation value. They applied this to 2.5m DEMs from the Star-3i system airborne X-band InSAR and the EagleScan Digital Airborne Topographical Imaging System (DATIS) LiDAR. The advantages of both rules combined yielded more accurate results. There was ~10% reduction in area error, >1m reduction in mean height error, and improved vertical accuracy in the fused DEM

Deng et al. (2011) combined a preferential rule and weighted fusion for merging SPOT-5, Radarsat-2 and COSMO-SkyMed DEMs. They preferentially selected elevations from intersecting pixels of the source DEMs based on prior knowledge. This was combined with an arithmetic fusion using correlation coefficients as weights. No gaps or void pixels in the fused DEM were observed, the accuracy was improved, and it showed more topographic detail. Perko and Zach (2016) adopted a global formulation for DEM fusion based on a convex energy functional that employed the Huber norm. The method was applied to 20cm DEMs extracted from stereo-pairs of UltraCam-X images, 1m DEMs generated from Pléiades tri-stereo satellite acquisitions and 1m stereo DEMs generated from TerraSAR-X Staring Spotlight images. The fusion generated a smoother solution with richer detail. The quality of the fused DEM also surpassed that of other fused DEMs produced by local methods.

Dong et al. (2018) used an edge-preserving guided filter for InSAR DEM fusion that combined the advantages of multi-orbit and multi-baseline InSAR observations. They applied it to 0.4 arcsec (~12m) bi-static TanDEM-X InSAR ascending/descending pairs with different normal baselines. The TanDEM-X RMSE was reduced by 6-16% in three sets of fusion experiments. Karakasis et al. (2014) implemented a forward transformation of input DEMs using five different spectral expansion methods (Chebyshev, Legendre, Tchebichef, Fourier and Cosine). This was followed by filter-based (low-pass and high-pass) fusion of their spectral/frequency coefficients. Finally, an inverse spectral transform was applied to return the fused coefficient set to the spatial domain. The method was applied to the fusion of 90m SRTM and 30m ASTER GDEM. A reduction in the mean absolute error (MAE) was achieved from 5.94m in SRTM and 6.02m in ASTER to as low as: 5.66m (Chebyshev), 5.61m (Legendre), 5.64m (Tchebichef), 5.85m (Fourier) and 5.66m (Cosine). They also experimented with a weighted version by combining all spectral components of input DEMs and this yielded a better performance than the filter-based method.

A linear combination approach for weight estimation using the variance and correlation of errors for sites with reference data was adopted by Pham et al. (2018). They exploited the relationship between slopes and weights in regions with reference data, then applied the developed relationship to other regions without reference data. The method was applied to 30m C-band SRTM and 30m ASTER GDEM v2. They also proposed an approach for constructing weight maps that uses K-nearest neighbour (K-NN) for grouping slopes and errors together to determine the local optimum weight. In Papasaika et al. (2009), fusion was carried out by means of a mathematical approach using weights. An adaptive thresholding was used to determine areas to be fused. They applied it to a 4m DEM produced by image matching of IKONOS satellite images and 2m DEM produced by airborne LiDAR scanning. No artefacts were observed in the fused DEM. Also, the fused DEM had the highest resolution and was temporally up to date.

Guan et al. (2020) fused the 30m SRTM1, 90m TanDEM-X and 30m AW3D-30 DEMs using a combination of sparse representation strategy and an adaptive regularisation variation model. This led to more realistic terrain texture information in the fused DEM. The sparse representation enabled a full extraction of terrain information from the low-resolution and high-precision 90m TanDEM-X DEM. The adaptive regularisation variation framework solved the terrain discontinuities (e.g., noise) and preserved topographical features, thus ensuring the fused DEM was close to the reality of the terrain. In a Q-Q plot, the errors of the fused DEM were closer to the line of best fit and the error variation showed a normal distribution. This indicated a good level of stability in the quality of the fused DEM.

### 4.4 Pros and cons

Table 4 summarises the pros and cons of the investigated methods. The simple and weighted average fusion methods stand out as the simplest methods in terms of implementation. However, simple averaging cannot adequately resolve artefacts and edge effects. These effects could manifest as surface discontinuities and artificial landform patterns in the fused DEMs. Also, correct height values in the source DEMs might be wrongly alternated. Despite these drawbacks, simple and weighted averaging are still preferred due to their low computational expense and fast processing speed. Variational models (e.g., TV-$L_1$, TGV-$L_1$ and Huber) can improve the fusion accuracy beyond the classic weighted averaging. This merit could come at the cost of additional computational expense and slower processing speed. Variational models are particularly good for morphologically complex landscapes or landscapes with high frequency contents such as high-density urban areas.

Super-resolution (SR) methods are good for noise suppression and integration of high-resolution and low-resolution information. With SR frameworks, more comprehensive and accurate terrain details can be achieved. With self-consistency measures, the relationship between the self-consistency distribution and the geospatial error distribution is not isomorphic. (i.e., they do not exactly correspond to each other). Even with the improvements in fusion quality, a more compressive formulation would be required for a precise prediction of the statistics of the geospatial errors from the self-consistency distribution. The maximum likelihood fusion scheme can improve the depiction of landforms in the fused DEM. However, it deals with each pixel individually and does not consider the spatial correlation in the fused images' pixels. For this reason, the fused DEM does not consider natural characteristics, such as smoothness, or other terrain representations.

The k-means clustering approach is data-driven and does not require any *a priori* information of the input DEM error. Also, only a few user-defined data distribution parameters are required for establishing/setting thresholds. However, it could lead to uncertainties due to the implicit assumption that more clustered elevations are more accurate. The multiscale Kalman smoother (MKS) algorithm can achieve dramatic improvements in the fused DEM, especially in steep channels where InSAR datasets are affected by layover and foreshortening.

Table 4: Pros and cons of the investigated methods

| DEM Fusion Method | Pros and Cons |
|---|---|
| Simple averaging | • Simple and easy to implement.<br>• Could lead to surface discontinuities and artificial landform patterns especially along the edges or blend zones. |
| Weighted sum of data with geomorphologic enhancement | • Less pretentious pre-processing.<br>• Enhanced technological connections while pre-processing.<br>• More transparent quality tracking and ease of update with additional data sources.<br>• This method is convenient due to its foundation on map algebra. |
| Weighted averaging | • Ease of implementation.<br>• Seamless transition in overlap zones.<br>• Minimisation of edge artefacts.<br>• Significant improvement in the representation of terrain details.<br>• If grids in the same class of terrain surface are allocated the same weight, the cost of fusion is reduced.<br>• Improved vertical accuracy.<br>• Reduction in invalid/void pixels.<br>• Combined use of weighted average and filtering algorithms could improve accuracy.<br>• Artefacts and discontinuities could be introduced.<br>• Could lead to loss of important terrain details.<br>• Already correct height values may be wrongly alternated. |
| Variational models | • Variational models can improve the fusion results beyond the classic weighted averaging.<br>• Reduction in noise effects.<br>• Better representation of building footprints.<br>• A weighted version of variational models can improve quality.<br>• It could increase the quality of fused DEMs especially in morphologically complex landscapes and areas with high-frequency contents such as built-up/urban areas.<br>• Some users might find it computationally intensive. |
| Sparse representations | • Computationally efficient, robust and flexible.<br>• Visual quality of original reconstructed signals can be retained. |
| supported by weight maps | • It accounts for prior information on terrain shapes (in a data dictionary), and the input DEM accuracies (supported by weights).<br>• In some cases, the quality of the fused DEM might be degraded. |
| Weighted average fusion with slope mask | • Local features are well captured.<br>• Fused DEM reserves long-term accuracy and inherits the high frequency spatial features from the source DEM.<br>• Improved accuracy |
| Wavelet domain weighted averaging | • Minimisation of elevation error dispersion |
| Self-consistency measures | • Improvement in quality.<br>• It is entirely data-driven whereby the elevation is estimated solely from the multiple data sets.<br>• It has significant overhead in terms of computational cost/expense.<br>• For a precise prediction of the statistics of the geospatial errors from the self-consistency distribution, a more compressive formulation would be required. |
| Regularised super-resolution (SR) | • It removes the noise and seamlessly fuses the high resolution and low-resolution information.<br>• More comprehensive and accurate terrain details.<br>• It provides a more accurate bank line.<br>• Visually attractive contour lines and comprehensive terrain features.<br>• The fused DEM is continuous and enhanced with details. |
| Multi-scale decomposition and a slope position-based linear regression | • Texture and detail information are clear and there is no mutation region.<br>• Effective noise detection and minimisation<br>• Consistency of trend signals. |
| Multiple-point geostatistical simulation | • The improved FILTERSIM method can achieve greater accuracy and better retention of terrain spatial structure than the FILTERSIM method. |
| Maximum likelihood fusion scheme | • Landforms are well depicted.<br>• The terrain smoothness, or other terrain representations are not effectively handled. |
| Linear combination approach for weight estimation | • The developed slope-weight relationship can be applied to other areas with similar geomorphology where reference data is not available. |
| MKS algorithm | • It does not require resampling of source data to the same resolution.<br>• Fused DEM exhibits smaller height uncertainty than in the source DEMs<br>• Dramatic improvements observed in the fused DEM, in the steep channels where shadowing and foreshortening could affect InSAR datasets.<br>• Improved mapping of channel cross-sections.<br>• Improvement in the accuracy of elevation estimates. |
| Extended Kalman Filter (EKF) | • Improvement in accuracy. |
| Guided filter | • It incorporates information on local spatial context leading to minimisation of the noise effect and data voids are filled automatically.<br>• It preserves terrain details through maintenance of gradient consistency and inclusion of terrain features.<br>• It is computationally efficient |
| Filter-based methods | • Improved vertical accuracy.<br>• Good conveyance of textural and edge content. |
| Fast Fourier Transform (FFT) | • Improvements in accuracy and completeness. |
| Convex energy functional (with β-Lipschitz continuous gradient) | • Generates a smoother solution, with more realistic terrain surfaces.<br>• Richer detail and higher quality.<br>• The smoothness of the resulting fused DEM can be defined by altering the regularisation parameters. |
| Combination of preferential rule and weighted fusion | • Elimination of gaps or empty pixels.<br>• Improved accuracy, with more topographic detail. |
| Combination of maximum rule and maximum correlation rule | • Improved vertical accuracy. |
| Clustering approaches (K-Means clustering) | • Only a few user-defined data distribution parameters are required for establishing /setting thresholds.<br>• Does not require any a priori information of the input DEM error.<br>• Possibility of uncertainty due to the implicit assumption that more clustered elevations are more accurate. |
| Bayesian inference | • Removal of spikes from buildings. |
| Adaptive regularisation variation model based on sparse representation | • More realistic terrain texture information.<br>• The sparse representation enables full extraction of terrain information.<br>• The adaptive regularisation variation framework solves the terrain discontinuities (e.g., noise) and preserves topographical features, thus ensuring the fused DEM is close to the reality of the terrain.<br>• Good level of stability in the quality of the fused DEM. |

### 4.5 Analysis of DEM sources and characteristics

Figure 12 shows the frequency of occurrence of the most common source DEMs in the reviewed studies. The most common is the TanDEM-X (TDX) DEM which was used 12 times in different studies. TDX is a product of the German Space Agency (DLR) and the data was acquired between 2010 and 2015 together with that of its twin satellite, TerraSAR-X (TSX). TDX is frequently made available by DLR at no cost to scientists for research purposes. Also, a considerable part of DEM fusion research was from Germany. SRTM and ASTER are also among the most frequently used DEMs, thanks to their global coverage and free availability. The first version of SRTM was released in early 2000 while ASTER version 1 (GDEM1) was released in 2009. Both DEMs have undergone several refinements over time and newer versions have been released since then. SRTM and ASTER are distributed in user-friendly formats, and this could partly explain their wide adoption even by less technical user groups all around the world. TSX was also a common source of elevation data used in combination with other datasets to generate DEMs through interferometry of TDX/TSX pairs (e.g., Du et al., 2019; Arief et al., 2020) and COSMO-SkyMed/TSX (e.g., Jiang et al., 2014). Some other DEMs that were used include ERS, Worldview, Cartosat and Radarsat. There was low usage of some of the SAR datasets such as ALOS PALSAR and COSMO-SkyMed. None of the DEMs from spaceborne LiDAR missions, e.g., ICESat and ICESat-2 were featured in the studies. Aside from their coarse spatial resolution, these space-borne LiDAR datasets are not traditionally distributed in the common GIS data formats and this might cause some reluctance among less technical user groups to adopt them.

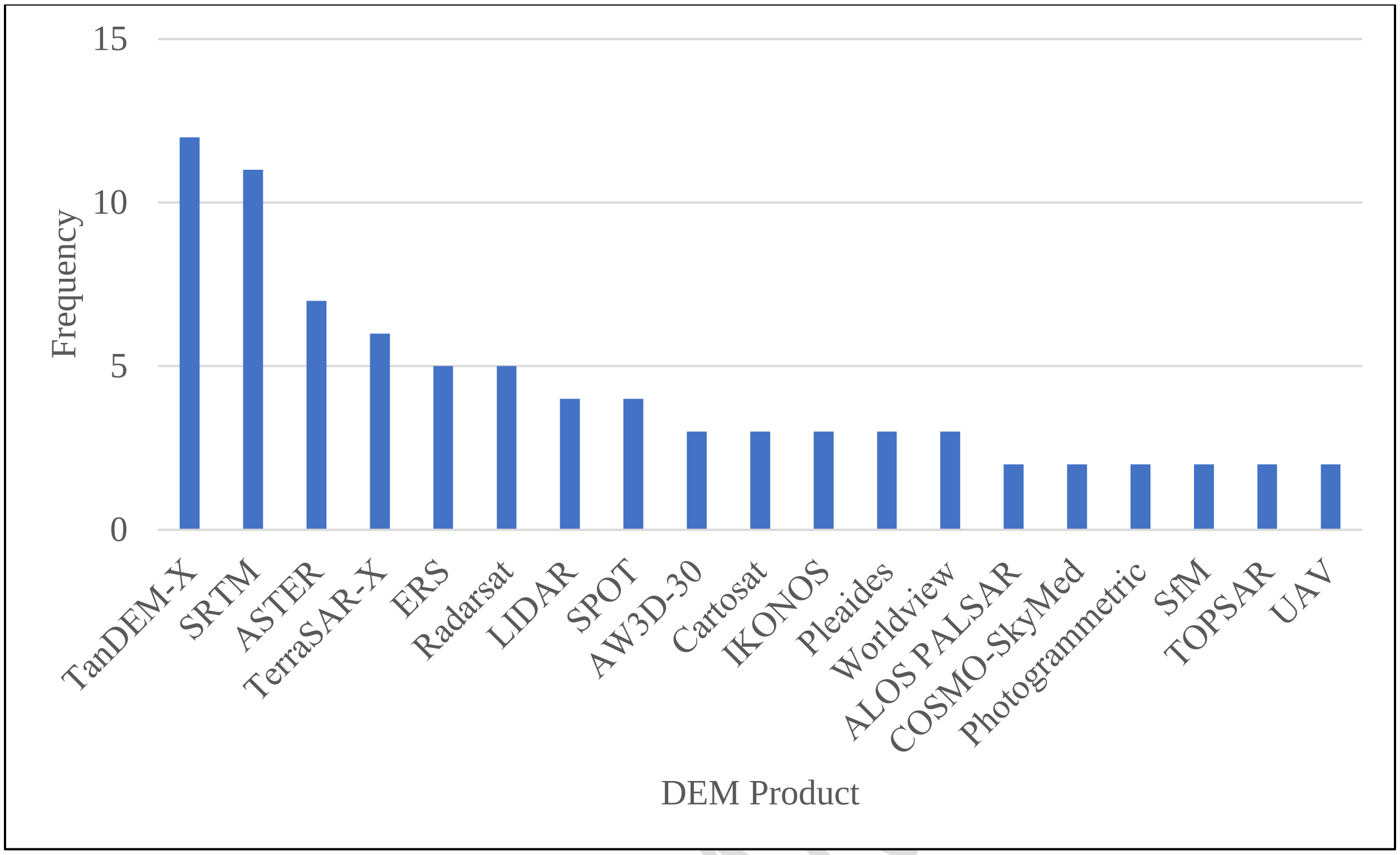

Figure 12: Frequency of occurrence of the most common source DEMs in the reviewed studies. Some DEMs are readily available to end-users at no cost (e.g., SRTM and ASTER) while others like the 12 m and 30 m TanDEM-X DEMs can be ordered by scientific users via a proposal submission.

Figure 13 shows the most frequently used reference DEMs/datasets in the reviewed studies. LiDAR DEMs (especially aerial LiDAR) were the most frequently used. High spatial resolution and small footprint LiDAR is a reliable validation dataset due to the typically higher accuracies achieved. Moreover, LiDAR can penetrate dense vegetation cover to realise the bare-earth topography. These attributes make it one of the most reliable datasets for validation. Next are ground control points (GCPs) and topographic maps/data. GCPs are commonly established by Global Navigation Satellite Systems/Global Positioning System (GNSS/GPS) surveys. However, GCPs are not available for most environments especially in difficult or challenging terrain such as mountainous regions. More so, the classical method of establishing GCPs is very laborious and time-consuming.

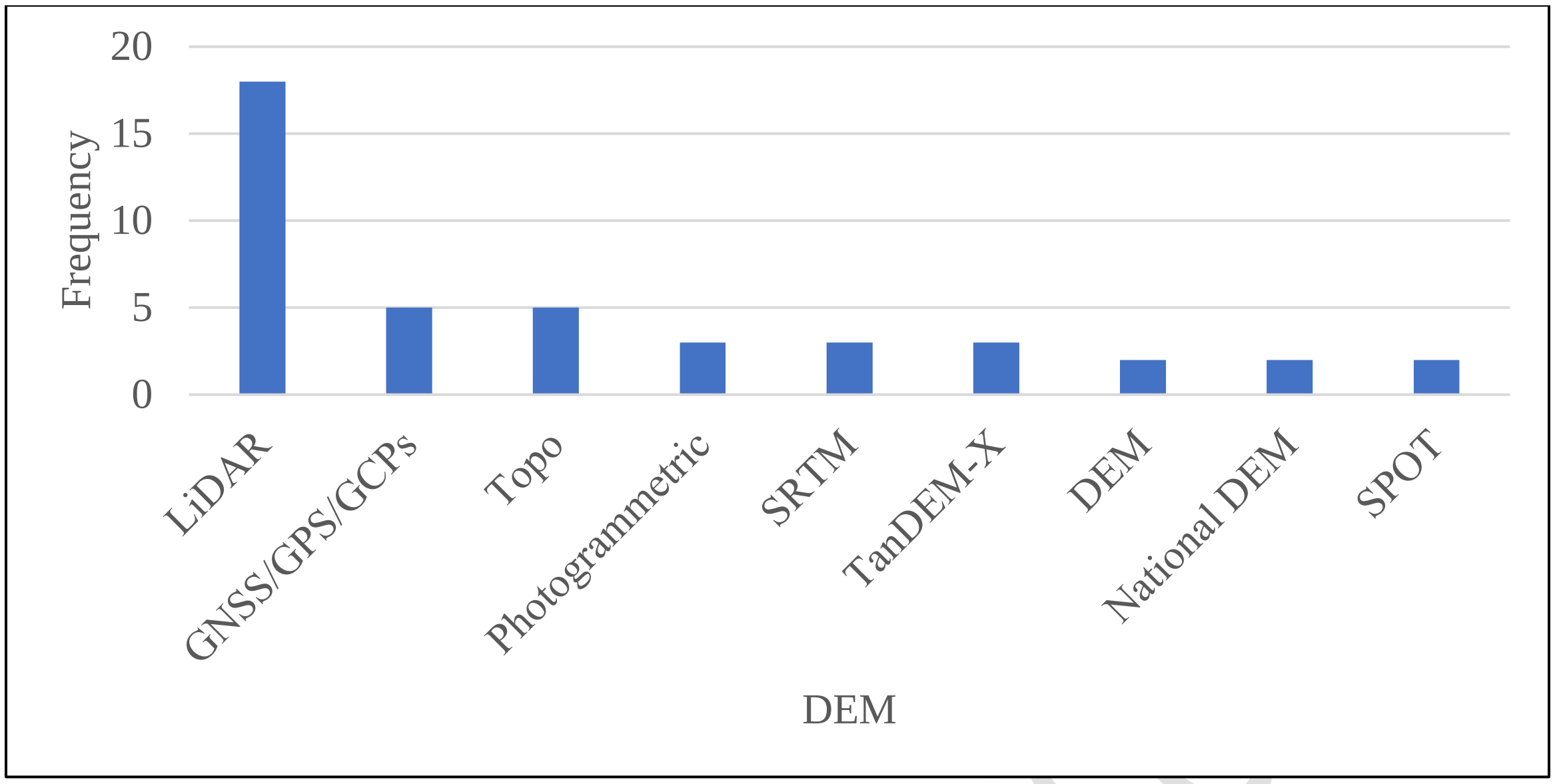

Figure 13: The most frequently used reference DEMs/datasets in the reviewed studies. Due to its highly achievable vertical and horizontal point positioning accuracy, airborne LiDAR is commonly adopted for the validation of fused DEMs.

Figure 14 shows the most common source DEM production technologies. Most of the source DEMs were generated through SAR interferometry (InSAR). This is corroborated by Figure 12 which shows that TanDEM-X, TerraSAR-X and SRTM were among the most common source DEMs. InSAR is closely followed by stereo-optical photogrammetry which is the method behind the ASTER GDEM. Other less commonly used methods include Structure-from-Motion (SfM) photogrammetry, image matching and stereo-radargrammetry. LiDAR technology is used in most of the reference DEMs for validation purposes (Figure 15).

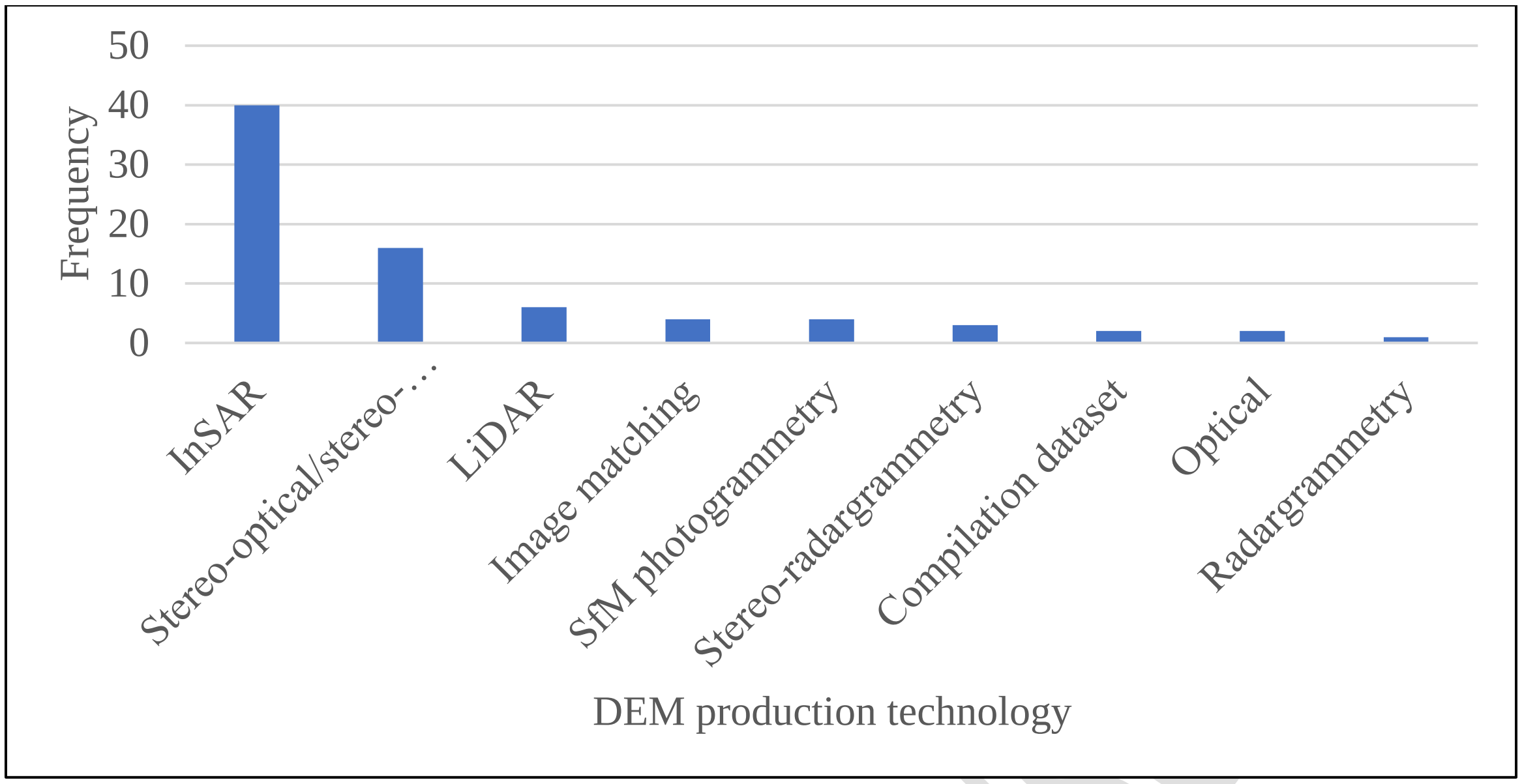

Figure 14: The most common source DEM production technologies featured in DEM fusion studies. For several years, InSAR has been used as an innovative technique for DEM generation.

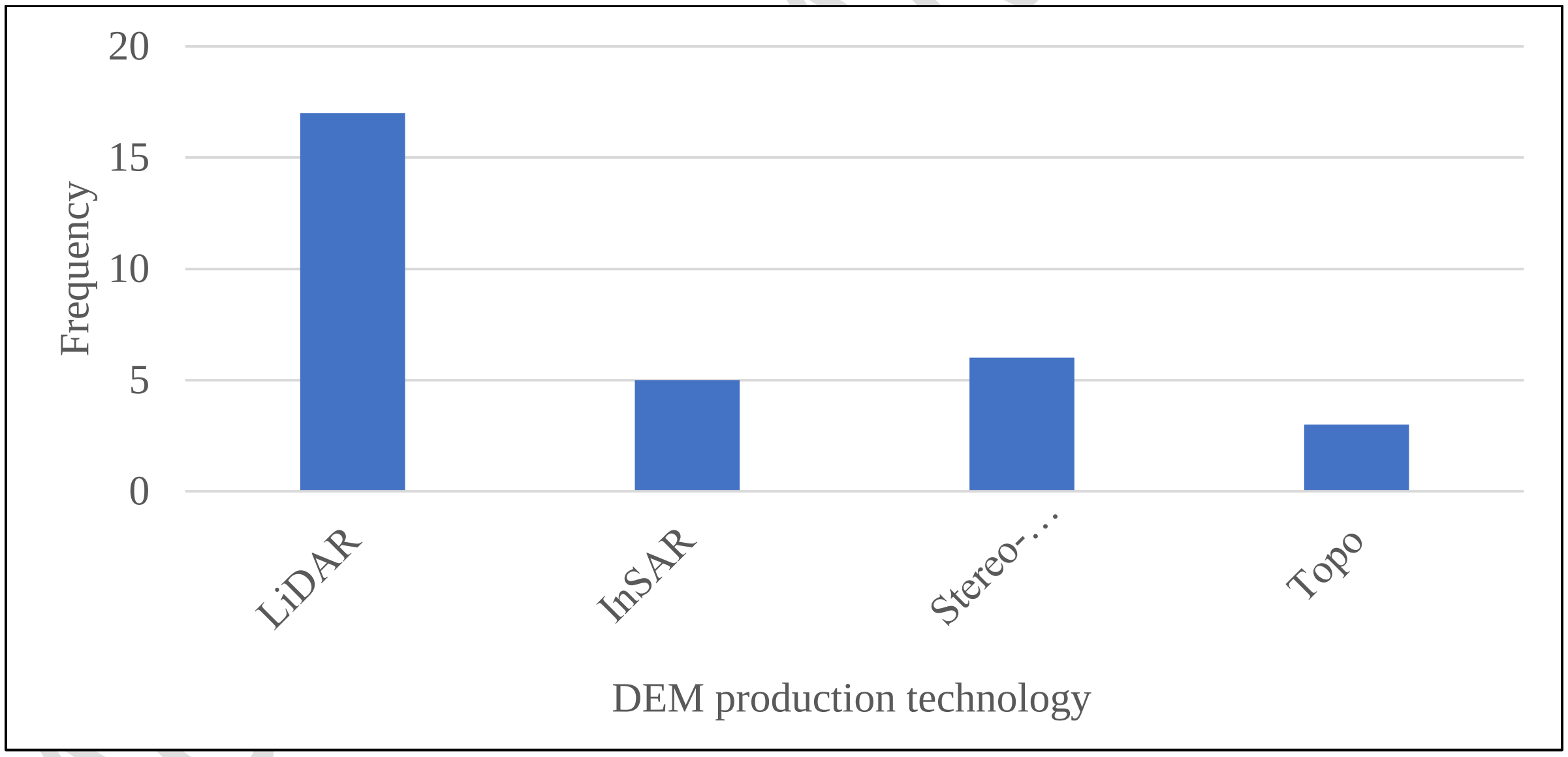

Figure 15: The most common reference DEM production technologies featured in DEM fusion studies. Small footprint airborne LiDAR provides very accurate elevation data and can effectively characterise the sub-canopy terrain elevation for validation of fused DEMs in densely vegetated terrain.

Figures 16 and 17 show the distribution of source and reference DEM spatial resolutions respectively. Most of the reference DEMs used for validation had a finer resolution than the source

DEMs being fused. A few DEMs with resolution >30m (e.g., ICESat LiDAR) were used as validation data.

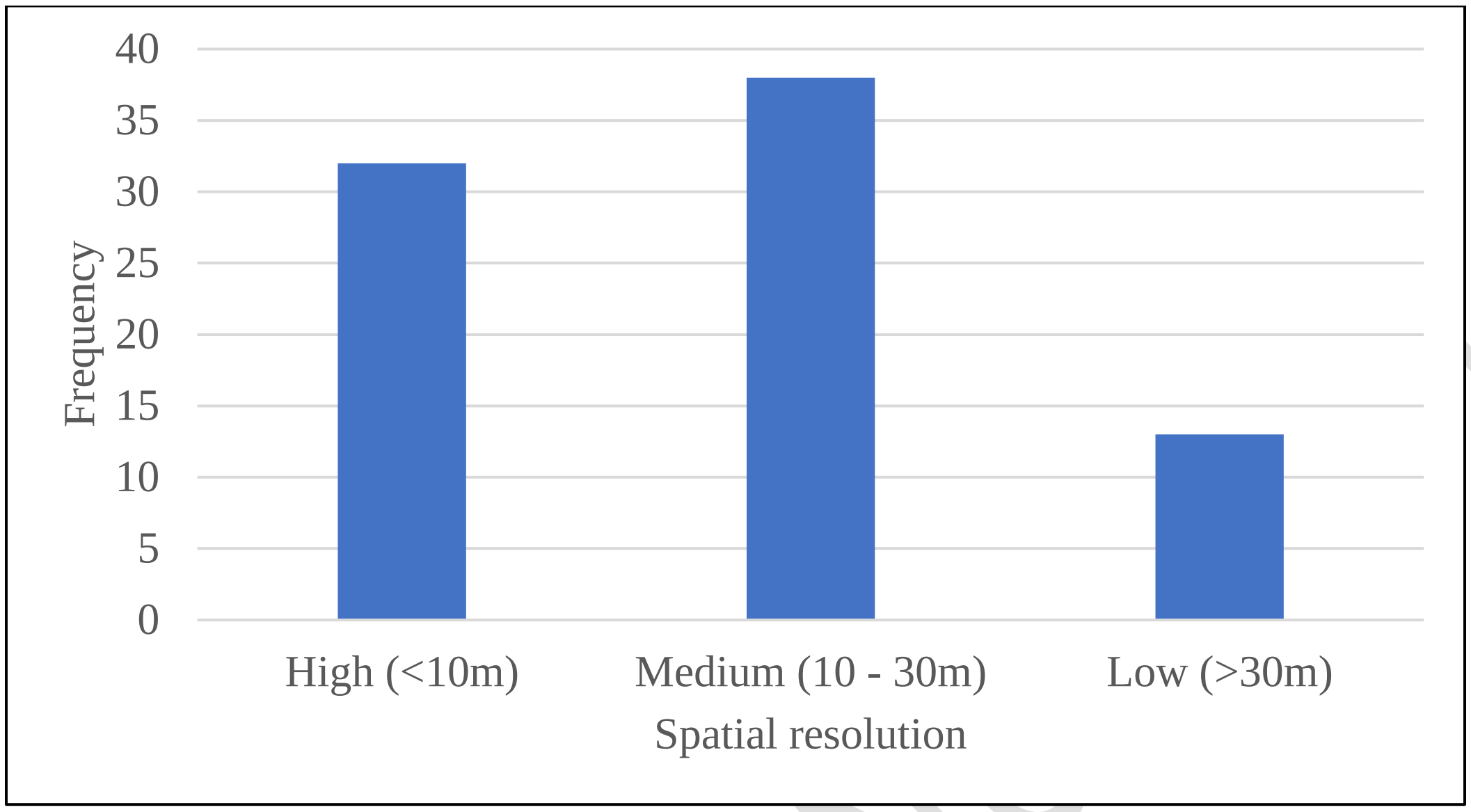

Figure 16: Distribution of source DEM spatial resolutions used in the investigated studies. Medium-resolution global DEMs such as the 30m SRTM and ASTER were commonly adopted, and this is likely due to their availability rather than their accuracy.

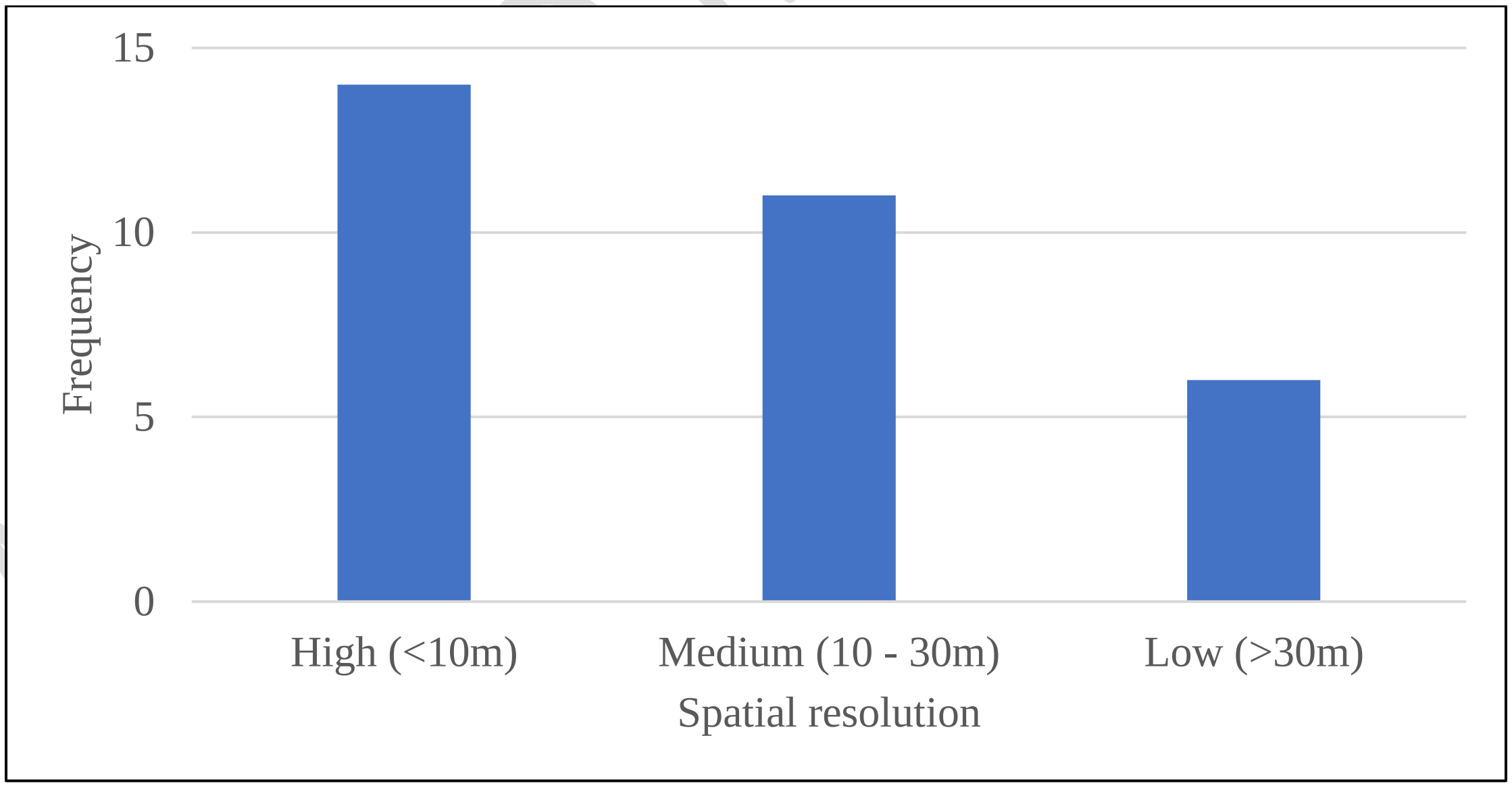

Figure 17: Distribution of reference DEM spatial resolutions used in the investigated studies. The spatial resolution of reference DEMs should exceed that of source DEMs since higher DEM resolution usually implies higher accuracy and more accurate representation of terrain details.

## 4.6 Parameters influencing weight maps and height error maps

Weight maps or height error maps (HEMs) are commonly used for modelling the height error distribution in source DEMs during fusion. Some DEM providers release HEMs along with the DEM products but not in all cases. For example, Bagheri et al. (2018) used the HEMs which were provided by the Integrated TanDEM-X Processor (ITP). In the HEM, a global uncertainty is usually reported for the DEM product, or in some cases, several standard deviations are reported (e.g., for different land cover or terrain classes) (Schindler et al., 2011). The global uncertainty is insufficient because differences in scene characteristics, processing methods, conditions under which the data was acquired, and differences in sensor characteristics imply varying accuracy in the DEM (Schindler et al., 2011). In WA studies, the fusion weights are usually based on specified conditions or HEMs which in turn are dependent on parameters such as interferometric coherence (Jain et al., 2014; Gelautz et al., 2003; Arief et al., 2020), distance from the transition line between source DEMs (Robinson et al., 2014), geomorphological characteristics (Schindler et al., 2011; Tran et al., 2014; Fu and Tsay, 2016), multiple search distances (Deng et al., 2019), or priority values of the most reliable InSAR elevations (Gruber et al., 2016).

Deo et al. (2015) noted that the HEM is dependent on factors that constitute the height of ambiguity of acquisition (range, perpendicular baseline and incidence angle), and that represent the outcome of rigorous error propagation in the determination of interferometric phase. However, the HEM is not always a reflection of the true quality of the input DEMs because it might not consider major error sources such as elevation offsets in the InSAR processing workflow and phase unwrapping errors (Deo et al., 2015). We have summarised the common parameters which are usually used for generating weight maps and HEMs and these are presented in Table 5.

Table 5: Parameters used for weight maps/height error maps or for modelling the height error distribution, and frequency of use

| Terrain parameter | No. of studies |
|---|---|
| Slope | 12 |
| Aspect | 5 |
| Roughness /entropy | 5 |
| Land use/land cover (LU/LC) | 2 |
| Interferometric coherence | 5 |
| Height of ambiguity | 6 |
| Edginess | 1 |
| Anisotropic coefficient of variation (ACV) | 2 |
| Variance of phase noise | 1 |
| Visibility | 1 |
| Topographic Ruggedness Index (TRI) | 1 |
| Topographic Position Index (TPI) | 4 |
| Ruggedness | 1 |
| Surface Roughness Factor (SRF) | 1 |
| Slope variability (SV) | 1 |
| Slope error | 1 |
| Standard deviation of elevation (SDE) | 2 |

In Fu and Tsay (2016), the same weight was applied to grids within the same class of terrain surface (slope and visibility) of mountains around Pingtung and Kaohsiung, Taiwan. In their method, the precision was not determined at every grid point, but weights were assigned to grids based on their class of terrain surface. Since grids in the same class of terrain surface were allocated the same weight, the cost of fusion was reduced. Tran et al. (2014) used a geomorphological approach for DEM fusion that was based on the accuracy assessment of SRTM and ASTER GDEM in flat areas, valleys and mountain slopes. Due to the varying topography and geomorphology, the stereo-optical technique used for generating ASTER and the InSAR technique used for SRTM give varying perspectives of the elevation. A weighted averaging fusion was applied in a topographic context, with a weighting scheme based on the MAE of slope and slope variability, including the landform type. The accuracy of the fused DEM was enhanced compared to the individual DEMs, and most artefacts were eliminated.

To ensure generalisability over the landscape, Bagheri et al. (2017, 2018) used training data of versatile land types such as inner-city areas (densely packed, relatively high buildings), residential areas (single-family homes and detached buildings), agricultural areas, and forested areas. From these areas, different kinds of spatial features describing landscaping and roughness properties of the land surface such as slope, aspect, surface roughness, and anisotropic coefficient of variation were extracted for training an artificial neural network (ANN). In addition, height residual maps were calculated from the corresponding DEM patches and LiDAR groundtruth data. The fusion based on ANN-predicted weights enhanced the quality of the DEMs with a relative accuracy of up to 50% in urban areas and 22% in non-urban areas. However, the HEM-based fusion did not exceed 20% in urban areas and 10% in non-urban areas respectively. Their results showed that the fused DEM based on ANN predicted weights had better quality than the original DEMs, and outperformed a more traditional fusion method using HEM-based weighted averaging.

Figure 18 shows the most common land use/land cover (LULC) classes at study sites covered in the studies while Figure 19 shows the percentage distribution of the most common terrain classes. The most common LULC classes covered were built-up/urban areas followed by forests. However, for most researchers, the choice of study site might be determined by data availability. In terms of the terrain class, 54% of sites were characterised by hilly/mountainous terrain or high relief. Mountainous terrain presents several challenges such as radar-specific imaging geometric distortions (e.g., foreshortening, layover and shadow) that occur in SAR images (Chen et al., 2018). This can have a negative impact on the quality of InSAR DEMs and their suitability for fusion. The analysis shows that 25% of sites were in flat lands or floodplains while 8% of sites were in areas of low relief or lowlands.

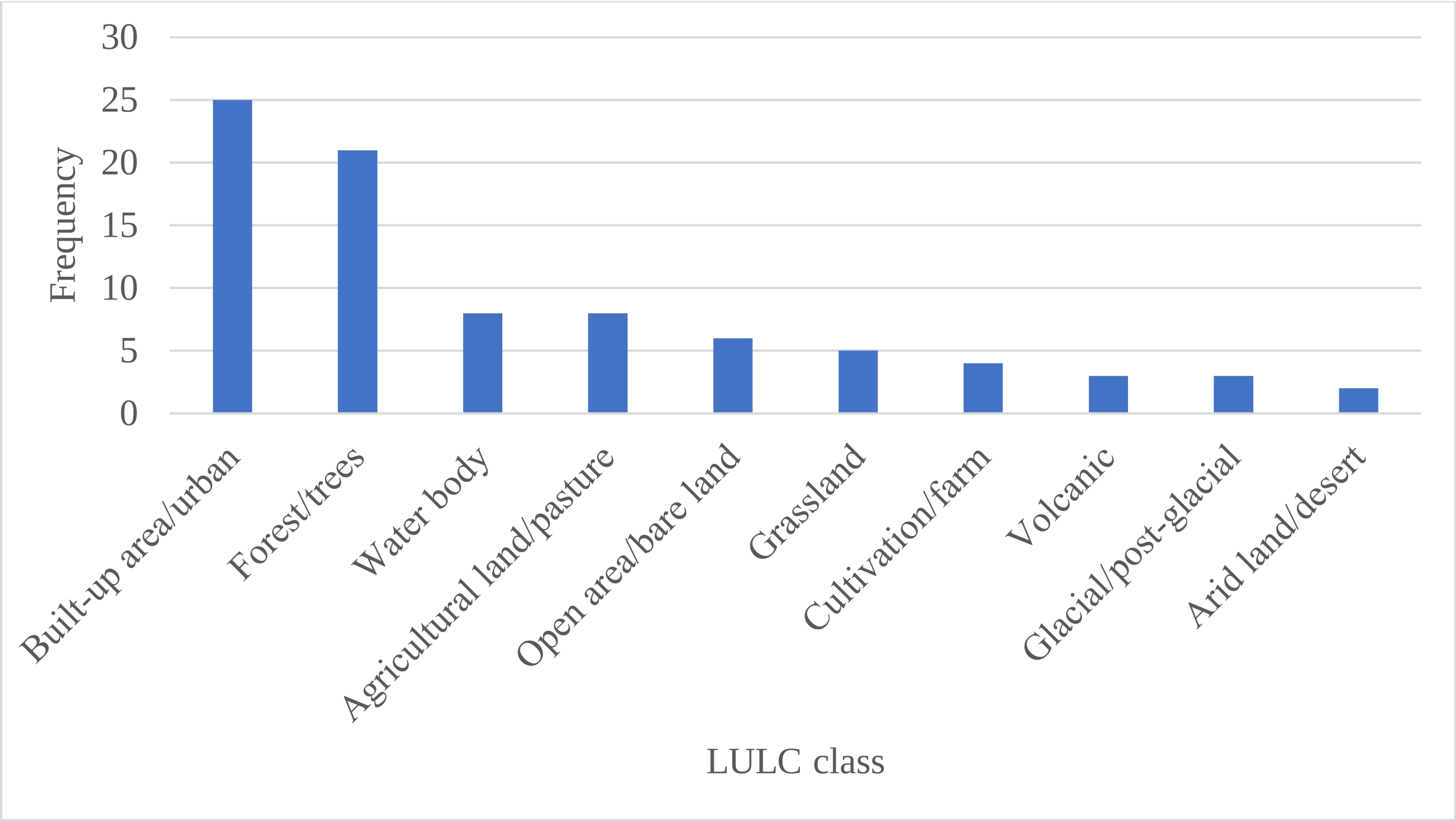

Figure 18: The most common land use/land cover classes at study sites covered by DEM fusion studies. Most sites were in built-up areas or forests. DEMs are known to exhibit local accuracy patterns in relation to the LULC distribution.

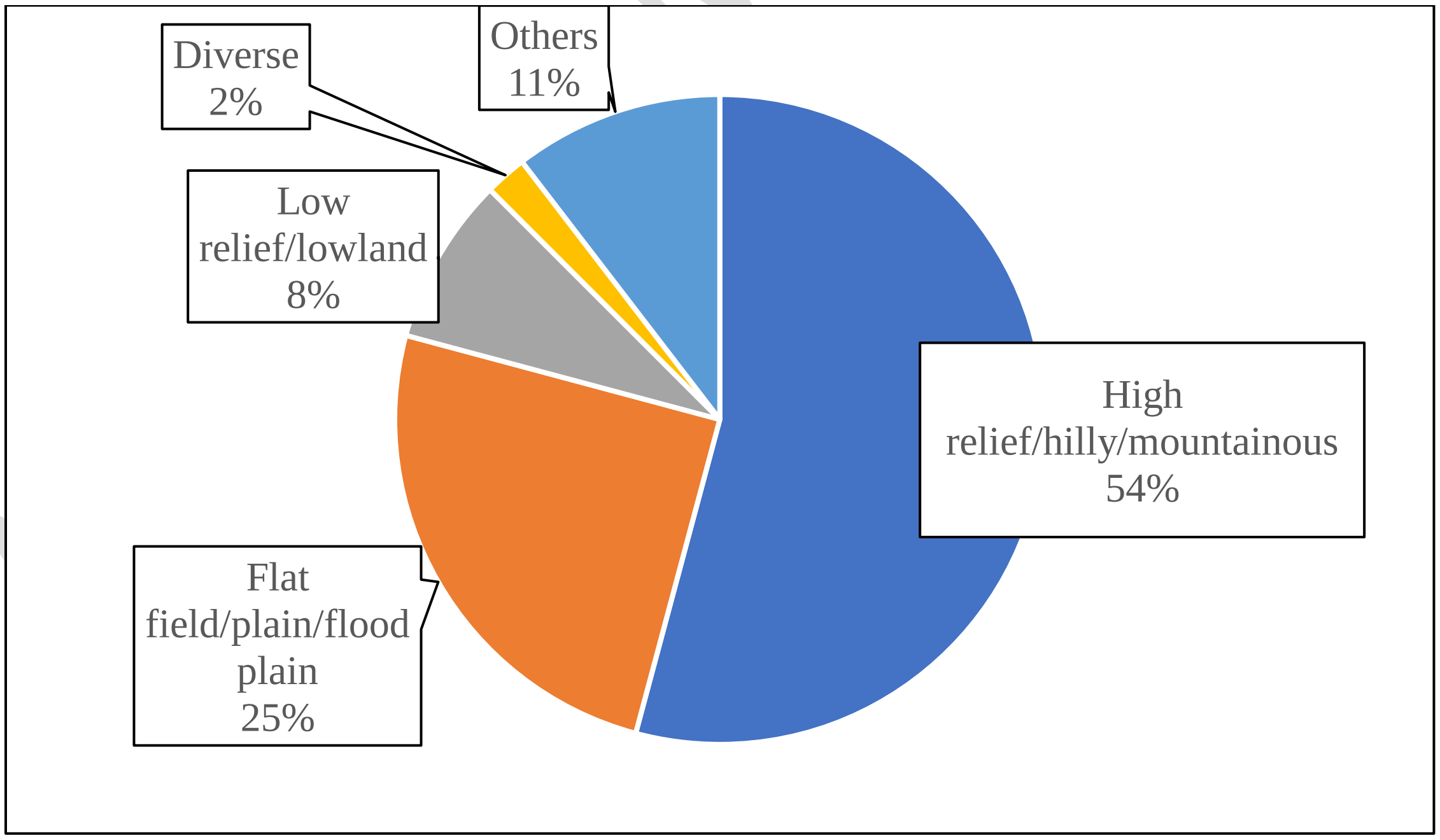

Figure 19: Percentage distribution of the most common terrain classes at study sites covered by DEM fusion studies.

## 4.7 Fusion of multi-geometry InSAR DEMs

The availability of many acquisitions in SAR flights is very beneficial. The different acquisitions have complementary characteristics which can be exploited to minimise phase unwrapping (PU) errors. Fusing ascending and descending InSAR DEMs compensates for the geometric distortions caused by data voids, especially in mountainous areas. For instance, InSAR acquisitions with larger heights of ambiguity (HoA) are more suitable for forested areas where the performance is strongly affected by volume decorrelation, whereas InSAR acquisitions that are processed with the dual-baseline phase unwrapping technique are more reliable with regard to phase unwrapping stability (Treuhaft and Siqueira, 2000; Gruber et al., 2016). Also, "an interferometric pair with long baseline and high frequency may have lower coherence, compared with that with shorter baseline and lower frequency, but it is more sensitive to the changes of ground elevation, particularly over vegetation areas" (Zhang et al., 2016). In rugged, mountainous and desert terrain, the DEM accuracy is limited by geometric distortions such as layover, shadow and foreshortening which affect the quality of the DEM (Gruber et al., 2016). In such areas, "crossing" acquisitions are acquired to improve the quality of the DEM with data from the opposite flight orbit direction (Borla Tridon et al., 2013 in Gruber et al., 2016). Fusion of multi-baseline or multi-frequency InSAR DEMs can resolve or minimise some of these errors (Zhang et al., 2016), and is a promising approach used in DEM fusion. According to Dong et al. (2021), the fusion of InSAR DEMs acquired from different orbital directions can also reduce random height errors and fill data voids caused by geometric distortions.

According to Gruber et al. (2016), the most relevant parameters that influence the reliability of heights in InSAR DEM acquisitions are the phase unwrapping method, the height error based on interferometric coherence, and the HoA. These parameters are also dependent on the processing method used and the InSAR orbit geometry (Gruber et al., 2016). Some previous studies have considered geometric parameters (e.g., local slope and aspect, baseline, incidence angle, satellite heading angle) as thresholds or to interpolate or mask out layover and shadow areas. However, using these geometric parameters could introduce additional artefacts (Deng et al., 2019). The most common approach for mosaicking InSAR DEMs is weighted averaging in which the weighted mean of all available height estimates is computed (Gruber et al., 2016). For example, the weights can be based on the interferometric height error which is used as a coherence-based measure for the quality of the InSAR DEM height with less reliable heights having lesser influence in the mosaicked DEM (Knöpfle et al., 1998; Gruber et al., 2016)

Bagheri et al. (2018) conducted three experiments to investigate the effect of fusing TanDEM-X raw DEMs with different properties. In the first experiment, two input DEMs with nearly similar properties (same orbit and look directions, and with similarities in incidence angles and HoAs) were selected. In the second and third experiments, the fusion was implemented on DEMs with different HoAs and baseline configurations. Three fusion methods were investigated in these tests – weighted averaging and variational models (TV-$L_1$ and Huber). They posited that a combination of ascending and descending pass DEMs improves the quality of the fused DEM in areas with difficult and complex terrain. Another benefit is the improvement of the DEM quality in areas affected by layover and shadow (Gruber et al., 2016; Bagheri et al., 2018)

### 4.8 Accuracy assessment

According to Hoja et al. (2006), the quality of satellite-derived DEMs over any given area is dependent on the terrain steepness and the land cover. Hence, it is beneficial to look beyond the single accuracy value given for the whole product and evaluate the accuracies of such DEMs on a regional basis. Absolute measures of elevation error do not give a complete picture of DEM quality (Hutchinson and Gallant, 2000, Papasaika et al., 2008). The most widely adopted approach for accuracy assessment is to use a reference height source such as GCPs or a DEM of higher accuracy. However, where reference data are unavailable or inadequate, non-classical measures may be adopted (e.g., comparison of elevations with surrounding neighbours, detection of local anomalies using slope and aspect, and inspecting histograms of elevation and aspect) (Papasaika et al. 2008). According to Schultz et al. (1999), a dense spread of groundtruth is required for a comprehensive evaluation and analysis of a dense array of elevation data generated from images. However, the resources to carry out such extensive groundtruthing campaigns are most times not feasible. The root mean square error (RMSE) is the most widely adopted statistical metric for assessing the vertical accuracy of DEMs. However, if the data does not follow a normal distribution, the RMSE could become unreliable (Fisher and Tate, 2006 in Fu and Tsay, 2016). Polidori and El Hage (2020), and Mesa-Mingorance and Ariza-López (2020) have compiled a review of DEM quality assessment methods and DEM accuracy assessment methods respectively.

In the reviewed studies, accuracy metrics have been reported for different experimental settings (e.g., fusion accuracy at sites with varying land cover and terrain characteristics, or of different dataset combinations) making it difficult to present an unbiased picture of the accuracy achievable by each method. The achieved accuracies are influenced by other factors such as the resolution characteristics of the source DEMs and the level of preprocessing carried out. Moreover, the authors did not utilise a consistent accuracy metric in their assessments. Since the RMSE was the most reported metric, the following summary is based on the average RMSEs. In each reviewed study, we selected the highest reported accuracy (lowest RMSE) and only the RMSE for one primary method were extracted from studies in which methods were compared. Summarily, researchers were able to achieve RMSEs below 2m with a few methods, e.g., Bayesian inference, linear combination approach for weight estimation, and multi-scale decomposition (with a slope position-based linear regression). With other methods such as weighted averaging, maximum likelihood, guided filter and variational models, authors were able to achieve RMSEs in the range of 2-5m. With the clustering approaches and simple averaging, RMSEs as high as 6.4m and 9.7m which are in the range of 5-10m were reported. The highest accuracies (lowest RMSEs) reported in the application of the Kalman filter, sparse representations supported by weights, fast fourier transforms (FFT) and the multiple-point geostatistical simulation (MPS) were greater than 10m.

### 4.9 Applications

The reviewed studies have shown the wide potential of DEM fusion for generating improved estimates of ground/surface heights and topography. Other applications include generation of wide-area DEMs, environmental hazard assessment, 3D building modelling, tactical military planning, flood and water flow modelling, surface water body detection, and other hydrological applications.

The weighted average method has been applied extensively in the generation of wide-area DEMs. The early versions of these DEMs usually suffer from many errors. Moreover, Schindler et al. (2011) note that such DEMs would have non-homogeneous data qualities as the properties in a given region would be identical to the input elevation sources for that region. The fusion usually involves the merger of both small-area and wide-area DEMs to produce global DEMs with added value. Schumann and colleagues (Schumann et al., 2016; Schumann and Bates, 2018) have argued that new technologies are not needed to meet the accuracy requirements of different scales of applications of DEMs but that existing DEMs should be merged through collaborative efforts to produce high-accuracy global DEMs. Examples of wide-area DEMs produced with application of weighted average fusion include TanDEM-X (Gruber et al., 2016), GTOPO30 (Gesch et al., 1999), the Global Multi-Resolution Terrain Elevation Data 2010 - GMTED2010 (Danielson and Gesch, 2011), EarthEnv-DEM90 (Robinson et al., 2014), and the European Union – Digital Elevation Model, EU-DEM (European Environment Agency, 2017; Mouratidis and Ampatzidis, 2019). These wide-area DEMs have been used in a wide scope of applications. For example, TanDEM-X (Figure 20) has been used for the measurement of pyroclastic flows (Albino et al., 2020), monitoring of subglacial volcanoes (Rossi et al., 2016), geomorphological mapping in high mountain environments (Pipaud et al., 2015), forest height estimation (Lei et al., 2021), forest structure mapping (Qi and Dubayah, 2016), bathymetric mapping (Zhang et al., 2016) and water reservoir storage estimation (Vanthof and Kelly, 2019). The EU-DEM (Figure 21) has also been applied in several studies such as: landform classification (Józsa and Kalmár, 2014) and extraction of geomorphological features (Mouratidis et al., 2017) while GTOP030 has been applied in the segmentation of physiographic features such as mountains, basins and piedmont slopes (Miliaresis and Argialas, 1999).

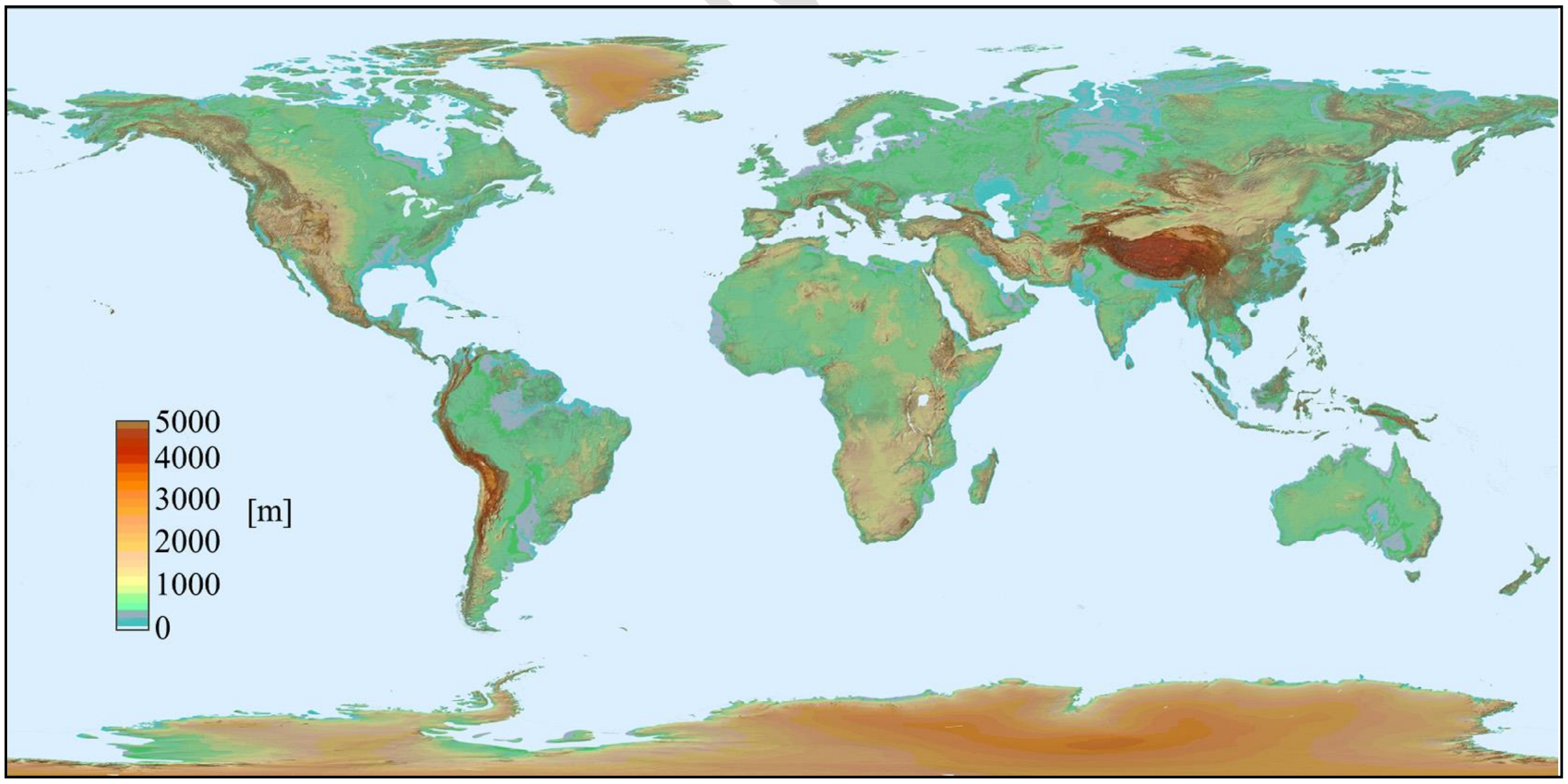

Figure 20: View of the global TanDEM-X DEM (source: Rizzoli et al., 2017)

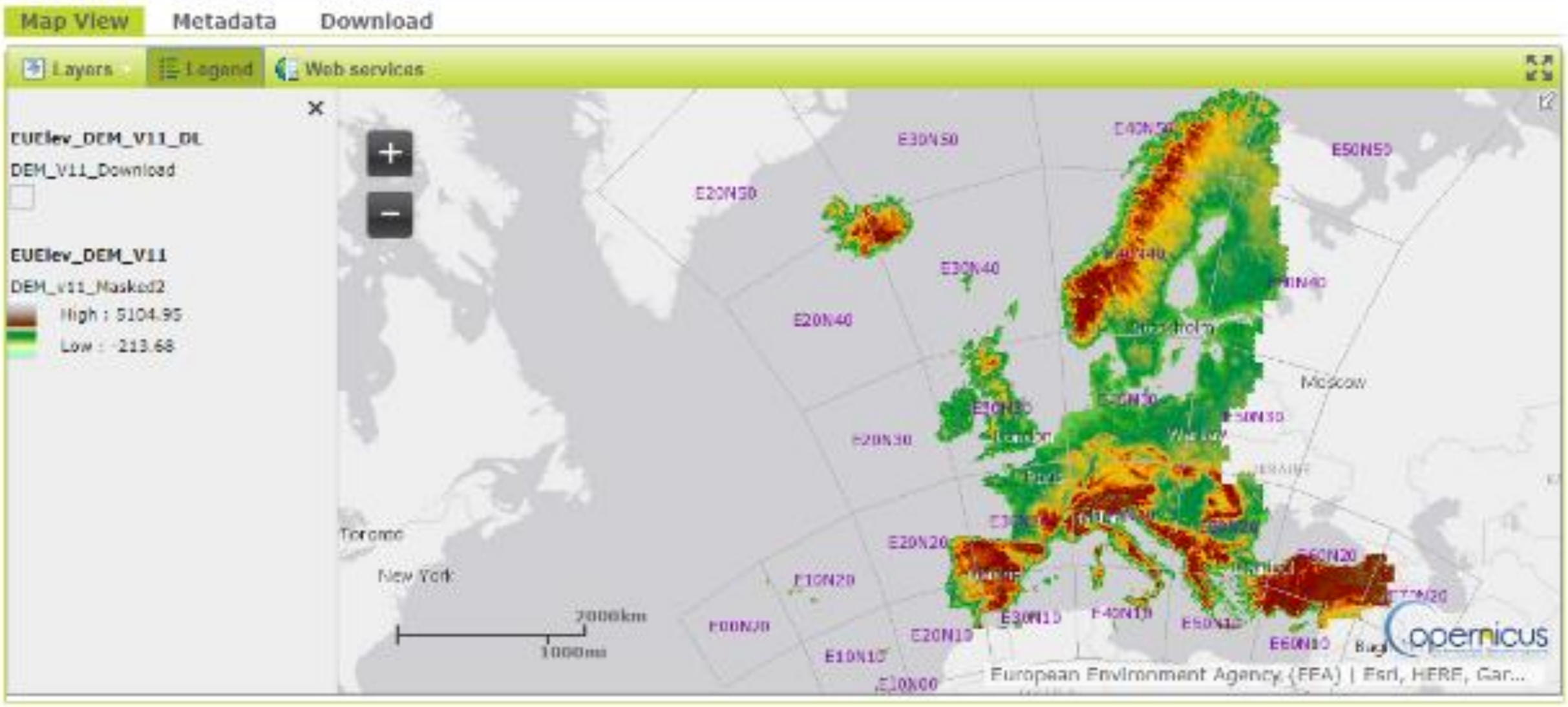

Figure 21: View of EU-DEM v1.1 (© Copernicus Programme)

Another application of fused DEMs is in volcano hazard assessment. For example, using Nevado del Ruiz volcano as an example, Deng et al. (2019) fused three DEMs from TanDEM-X, terrestrial radar interferometry (TRI), and Structure-from-Motion (SfM) to derive a new 10m DEM. They compared simulated inundation zones obtained from the new DEM and found significant differences (e.g., longer lahar run-out distance, highly channelized flows in a river channel and larger extent of thicker deposits) with respect to the 30m SRTM DEM. The fused DEM enabled a more detailed volcano hazard assessment.

Petrasova et al. (2017) explored the effects of DEM fusion on water flow modelling in two case studies drawn from the context of precision agriculture. The first application fused a LiDAR-based DEM with a fused set of UAV-acquired DEMs. In the second application, they fused a georeferenced, physical sand model continuously scanned by a Kinect sensor with a LiDAR-based DEM of a watershed to computationally simulate and test methods for controlling stormwater flow. The results of both experiments demonstrated the importance of robust and seamless DEM fusion for the realistic simulation of water flow patterns.

Du et al. (2019) presented a novel approach for simultaneous surface water body detection and DEM generation using multi-geometry TanDEM-X pairs (ascending and descending orbits). Due to geometric distortions (e.g., layover and shadow) and the limitations of the experiential threshold-based methods, it is difficult to extract reliable and complete water body information from single-geometry SAR images (Hahmann et al., 2010; Hahmann and Wessel, 2010; Du et al., 2019). They generated the final water body map by fusing the single-geometry TSX/TDX pairs. In addition to the layover and shadow maps, a weighting scheme was designed using the interferometric phase error and perpendicular baseline. In their results, the fused DEM refined by the detected water bodies showed a satisfactory performance in separating water bodies from linear objects such as bridges and in calculating the elevations of water areas.

Bagheri et al. (2019) investigated the potential of level-of-detail (LOD) 1-based 3D building modelling from OpenStreetMap (OSM) and remote sensing data enhanced by multi-source and multi-modal DEM fusion. The DEMs were derived from Cartosat-1, TanDEM-X raw DEMs and stereo-optical photogrammetry of TerraSAR-X and Worldview-2 images. The DEM fusion pipeline included ANN-based weighted averaging for Cartosat-1 and TanDEM-X, while variational models (TV-$L_1$ and Huber) and weighted averaging were used for fusing TanDEM-X raw DEMs. Their findings showed that the height information from the fused DEMs was more reliable than the original data sources. Secondly, the study confirmed that simple, prismatic building models could be reconstructed through the combination of building footprints with elevation data derived through DEM fusion. Figure 22 shows the LOD1 3D reconstruction results consisting of prismatic building models generated by combining the height information derived from the fused DEMs and OSM-derived building footprints. Generally, all the models were systematically biased when compared to a model produced from high-resolution LiDAR data. The LOD1 reconstruction results along with original building outlines from OSM were quantitatively evaluated by comparison with the high-resolution LiDAR model. There were improvements in the RMSEs of the reconstructed models from 10.01m in Cartosat-1 and 10.16m in TanDEM-X to 9.97m in the ANN-based fusion of Cartosat-1 and TanDEM-X, 9.5m in the weighted average fusion of TanDEM-X, 8.95m in the TV-$L_1$ fusion of TanDEM-X, and 9m in the Huber-based fusion of TanDEM-X. These results demonstrate the application of DEM fusion for wide-area prismatic building model generation (at LOD1 level).

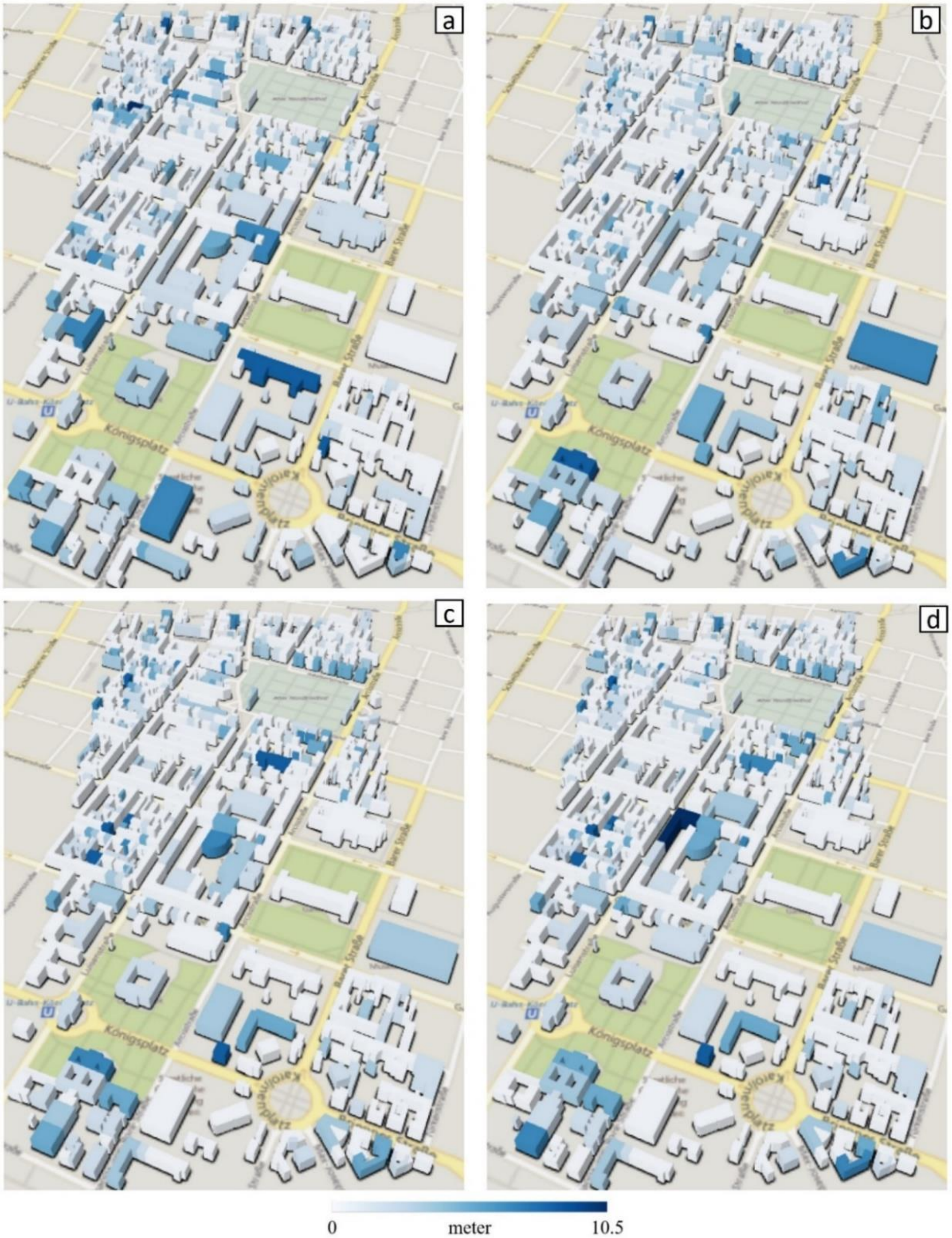

Figure 22: LOD1 3D prismatic building models in Munich, Germany generated by Bagheri et al. (2019) through the combination of OSM-derived buildings with, (a) ANN-based fusion of Cartosat-1 and TanDEM-X (b) WA-based fusion of TanDEM-X raw DEMs (c) TV-$L_1$ fusion of TanDEM-X raw DEMs, and (d) Huber-based fusion of TanDEM-X raw DEMs

Chaabane (2008) employed DEM fusion to improve topographic profile accuracy. The coherence information was introduced in the fusion process to reflect the reliability of the elevations. Three conjunctive operators were used - coherence weighted average, maximum coherence and coherence mixture decision. In the outcome, the resulting DEMs were more accurate than the sources, and this justified the fusion process.

Slatton et al. (2002) merged space-borne InSAR data from the ERS-1/2 platforms with airborne C-band InSAR data from the NASA/JPL TOPSAR platform to estimate high-resolution topography of the Finke River Gorge in Central Australia. The fusion was implemented using the MKS framework. They observed that the fused DEM provided the most dramatic improvement in the steep channels where shadowing and foreshortening could be problematic. The fused DEM also enabled spatio-temporal evaluation of changes in the fluvial structure of the river channel and monitoring of structural changes in the meandering channel portions.

## 5. Challenges, Open Issues and Future Directions

It is crucial to find a balance between end-user expectations and the capability of the fusion methodology. Other considerations include the robustness of the fusion method, its computational efficiency and global convergence (Papasaika et al., 2009). For example, a high-quality fusion method that is very computationally intensive might not be of value to users who are limited by computing resources. Also, an advanced fusion algorithm could be very computationally expensive but still not deliver the desired outcome. Thanks to the increased availability of global DEMs and computing resources, it is now easier to transform fusion experimentations into cost-effective packages. It is not only important to have access to DEMs and software, but also to workflows for fusion to develop information products. For this reason, it is essential to provide step-by-step procedures or recommended approaches that make use of readily available wide-area DEMs. These procedures can be made available in a growing variety of remote sensing and GIS software packages and programming languages such as R, Python and QGIS.

In most of the reviewed studies, a pre-analysis to determine the most relevant or influential terrain parameters for preparing weight maps was found lacking. A major contribution of this review is the detailed presentation of parameters used for preparing weight maps/height error maps to model the height error distribution in source DEMs. More so, since sites have different characteristics, the quality of fusion is largely dependent on this consideration. However, it is not mandatory to consider all possible parameters in each case as increasing the number might not necessarily improve the fusion quality. A pre-analysis is recommended whereby all available parameters are tested based on their correlation with the height error. Such pre-analysis can resolve multicollinearity, and this can lead to the elimination of less influential parameters. This review has also shown that the interest to screen the DEMs for non-ground components such as buildings and forests has been low, with only a few studies incorporating it in their pre-processing workflow.

Despite the development of more complex methods, the weighted averaging (WA) has remained the most common DEM fusion technique due to its simplicity and ease of implementation. Further research can develop improved WA schemes based on a more comprehensive set of terrain parameters, and with a wider sample of landscapes and geomorphological contexts. Going further, the recent advancements in machine learning/deep learning (ML/DL) are another avenue to key into for future research. When applied to remote sensing data fusion, ML algorithms present several advantages. For example, they can model complex class signatures, do not make assumptions on the nature of the data distribution (i.e., are non-parametric), and can accept a wide variety of input predictor data (Maxwell et al., 2018). Statistical ML has also played a leading role in the development of new data fusion methods through the incorporation of correlation analysis, prior knowledge and entropy metric, non-linear relations, and inherent statistical structures of input data (Guan et al., 2018). When combined with remote sensing, ML can lead to improved automation of data analysis and the discovery of new insights from complex data (Lary et al., 2018; UK Parliament POST, 2020). The use and adaptation of neural networks in remote sensing image fusion is a very active area of interest. This is evident in a large yearly turnover of scholarly publications on the subject. The focus has largely been on the fusion of traditional digital images derived from a variety of sensors with little attention on digital elevation models. The literature survey of DEM fusion methods has not shown any recorded use case of ML/DL methods to the actual fusion of DEMs. However, modest gains have been recorded in the development of a neural network support framework for weight map prediction as a precursor to the actual DEM fusion process (for example, Bagheri *et al.,* 2017, 2018). Although modest gains were recorded, the solution did not integrate the fusion process into the neural network setup. Summarily, ML/DL has profound advantages for DEM fusion such as in the preparation of weight maps that are proportional to the expected height residuals, and that can cater to the complex characteristics of variable landscapes. The computation burden of some methods, technical complexity and processing time is also a limitation that affects the easy implementation. Although the developed fusion methods are laudable, there has hardly been any implementation to derive software tools or add-ons that are easy to use by end-users. Thus, researchers can key into ML/DL to automate either data-driven or process-driven DEM fusion processes (or a combination of both). The emergence of powerful and dedicated computers capable of large-scale parallel computation, as well as the development of statistical inference algorithms, have enhanced immensely, the capabilities of ML and AI-based computational techniques for extracting scientific knowledge and closing the gaps between theoretical models and practical implementations (Humphreys et al., 2020).

Decision-level fusion is a high-level information fusion that has not yet been fully explored for DEM fusion. It integrates the results or outcomes from multiple algorithms to yield a final fused decision (Zhang, 2010). It makes use of value-added data where the initial images are processed separately for information extraction (Pohl and van Genderen, 1998). With decision-level fusion, decision rules can be applied to source DEMs to resolve differences, reinforce the common interpretation, and furnish a more detailed understanding of the observed terrain.

The remote sensing community should take advantage of the readily available global multi-sensor DEMs to create value-added DEMs through fusion. Value-added DEMs produced through fusion present numerous strategic benefits especially regarding Geospatial Data Infrastructures (GSDIs) and the United Nations Sustainable Developmental Goals (SDGs). Such fused DEMs are:

- crucial datasets for integration in models that deal with environmental changes and hydrological processes in the earth and marine ecosystems and in studies dealing with the physics and chemistry of the earth;
- important for infrastructural transformation through improved topographic maps and digital elevation data for urban mapping, engineering works and upgrade to the research expertise of present and future researchers;
- important for modelling topographic processes that govern the functioning of the biosphere and the climate system, water cycling and climate change;
- important datasets for biodiversity and marine habitat conservation;
- key parameters in models dealing with ice sheet mass balance;

- key inputs for habitat mapping, and environmental modelling for food and water security, climate change mitigation and adaptation; and for enhancing the resilience of plants and wildlife to biodiversity changes;
- crucial for providing improved estimates of the vertical structure of forests, above-ground biomass, and for forest canopy modelling.

Recent satellite LiDAR missions such as the recently launched ICESat-2 and GEDI provide global height distributions and are particularly beneficial in forested regions. Although the working principle of LiDAR is similar to Radar, the former operates in shorter wavelengths and has less beam divergence (Gerck and Hurtak, 1992 in Iqbal, 2010). These qualities enable LiDAR to capture local landscape variations better and conduct regionally focused studies on a reliable and short-term basis (Iqbal, 2010). However, there is a strong reluctance to use LiDAR data by some researchers because of its complexity regarding the pre-processing requirements and unfamiliar formats. These human capacity and technical requirements as well as the storage requirements of the data put it out of reach for most non-technical user groups in global south research institutes. Hence, international collaborations can overcome these challenges by exploiting the vast coverages of optical, InSAR and LiDAR global DEMs to provide analysis-ready fused DEMs on a national and regional scale. Such datasets can be demand-driven and standardised into tiles for easy distribution.

Most of the readily available global DEMs were produced with disparate techniques, are heterogeneous, and come in coarse resolutions. Consequently, wide-area DEM fusion is critical if digital elevation data is to be fully assimilated into the spatial data infrastructures of data-sparse regions of the world especially the low-income countries of Africa and the global south. It however requires significant technical expertise, and this might pose a problem for such parts of the globe where adaptive capacity is a challenge. Some initiatives have tried to shorten the skills gaps in the use of earth observation data e.g., AfriSAR and SAR-4-Africa. There has been a steady increase in the number of public wide-area DEMs (e.g., SRTM, ASTER, AW3D, NASADEM). However, these public DEMs are limited by their sampling resolution which results in an inadequacy of high-quality reference/validation data, ground-truth and training samples. Moreover, a variety of fusion methods with different performances have been proposed by researchers. It is pertinent to ensure a fair and transparent comparison of these datasets and methods. Hence, the need for the creation and testing of benchmark datasets for assessing the quality of DEM fusion. Organisations such as the International Society for Photogrammetry and Remote Sensing (ISPRS), IEEE GRSS and European Spatial Data Research (EuroSDR) have supported the organisation of benchmarks. Benchmarking allows for the exchange of thoughts and theories, independent research and joint experiments, often leading to universal conclusions (Bakula et al., 2019).

The vertical accuracy and some aspects of the terrain height representation have been addressed by several of the reviewed studies. Other desirable attributes include topological consistency (morphological structures), terrain continuity and completeness, and these can be covered in the future by other researchers. Notably, the fusion of DEMs in difficult terrain such as volcanic regions and arctic regions with ice sheets has not been fully explored.

With the increase of space missions and satellites being launched to other planetary bodies, new perspectives of the topography of the planetary bodies are being realised. Already, DEMs are being produced to depict the topography, geomorphology and hydrology of these planets. For example, the Mars MGS MOLA DEM 463m v2 is based on data from the Mars Orbiter Laser Altimeter (MOLA; Smith et al., 2001). A blended product known as Mars MGS MOLA - MEX HRSC Blended DEM Global 200m v2 is a blend of elevation data derived from the Mars Orbiter Laser Altimeter (MOLA) and the High-Resolution Stereo Camera (HRSC) (Fergason et al., 2018). More recently, the HiRISE camera onboard the Mars Reconnaissance Orbiter (MRO) has captured thousands of stereo pairs, enabling the creation of high-resolution DEMs (Hepburn et al., 2019). However, vast areas of these planets are unexplored and with the increased interest in space exploration, and the increasing number of extra-terrestrial DEMs, fusion will become a viable solution for generating more comprehensive topographic maps of the planets. The increasing interest in outer space exploration means that the fusion of extra-terrestrial DEMs is likely to become an important area of research in the future.

**ACKNOWLEDGEMENTS**
The authors are grateful to the University of Cape Town (UCT) Management and UCT Libraries for sustaining remote access to library resources during the COVID-19 lockdown. We also appreciate Dianne Steel and Tamzyn Suliaman (University of Cape Town Libraries) for their invaluable assistance with the systematic review approach and library resources. Special thanks to Dr. Matthew Page from Monash University Australia (co-lead of the 2020 update of the PRISMA guidelines) for his helpful feedback on our review methodology. We appreciate the following colleagues for their helpful comments and insightful feedback - Lisah Ligono (Dept. of Geomatics Engineering & Geospatial Information Science, Jomo Kenyatta University of Agriculture and Technology, Kenya), Michael Orji and Olagoke Daramola (Dept. of Surveying & Geoinformatics, University of Lagos, Nigeria), Assoc. Prof. Dr. Arif Oğuz Altunel (Department of Forest Engineering, Kastamonu Üniversitesi, Turkey), Ikenna Arungwa (Dept. of Surveying & Geoinformatics, Federal University of Technology Owerri, Nigeria), Caleb Ogbeta (Dept. of Surveying & Geoinformatics, Bells University of Technology, Nigeria) and Johanson Onyegbula (Dept of Natural Resources & Environmental Science, University of Nevada, Reno, USA). We also thank Samuel Akinnusi, Erom Mbu-Ogar, Waliyah Adedokun, Hamed Olanrewaju, Abdullahi Hamzat, Imole Okediji and Andy Egogo-Stanley (Department of Surveying and Geoinformatics, University of Lagos) for their assistance with some data arrangement and indexing tasks, computer-aided plots and visualisations, and helpful comments. Special thanks to academic staff at the Department of Surveying and Geoinformatics (University of Lagos) and the Geomatics Division (University of Cape Town) for their support. The Covidence software Product/Support team generously granted us a free license to use their software. Lastly, we are grateful to the Editors and anonymous reviewers for their painstaking multiple reviews which improved the quality of the manuscript.

## FUNDING

CJO received funding support through the University of Cape Town's Postgraduate Funding Office (UCT PGFO) International Student Award for Doctoral studies.

## AUTHOR CONTRIBUTIONS

**CJO** – Conceptualisation, writing (original draft, review and editing), methodology, data curation, formal analysis and investigation, visualisation, validation; **JLS** – Conceptualisation, Writing (review and editing), methodology, validation, resources, supervision.